\documentclass[11pt]{article}

\usepackage[preprint]{acl}

\usepackage{microtype}
\usepackage{graphicx}
\graphicspath{{latex/}}
\usepackage{subcaption}
\usepackage{booktabs} 
\usepackage{multirow}
\usepackage{xcolor}
\usepackage{colortbl}
\usepackage[most]{tcolorbox}
\usepackage{listings}
\usepackage{enumitem}
\usepackage{placeins}
\tcbuselibrary{listings,skins,breakable}

\usepackage{hyperref}
\usepackage{amsmath}
\usepackage{amssymb}
\usepackage{mathtools}
\usepackage{amsthm}
\usepackage{algorithm}
\usepackage{algorithmic}
\usepackage{makecell}

\usepackage{caption}
\usepackage{seqsplit}

\usepackage[capitalize,noabbrev]{cleveref}

\theoremstyle{plain}

\theoremstyle{definition}

\theoremstyle{remark}

\usepackage{times}
\usepackage{latexsym}

\usepackage[T1]{fontenc}

\usepackage[utf8]{inputenc}

\usepackage{microtype}

\usepackage{inconsolata}

\usepackage{graphicx}

\title{Back to Basics: Improving Molecular Understanding in LLMs via SMILES–Graph Translation}

\author{
  Wenda Wang \\
  Gaoling School of Artificial\\
  Intelligence \\
  Renmin University of China \\
  Beijing, China \\
  \texttt{wangwenda87@ruc.edu.cn} \\
  \And
  Jinjia Feng \\
  Gaoling School of Artificial\\
  Intelligence \\
  Renmin University of China \\
  Beijing, China \\
  \texttt{jinjia\_feng@ruc.edu.cn} \\
  \And
  Zhewei Wei \\
  Gaoling School of Artificial\\
  Intelligence \\
  Renmin University of China \\
  Beijing, China \\
  \texttt{zhewei@ruc.edu.cn} \\
}

\begin{document}
\maketitle
\begin{abstract}
Recent advances in molecular large language models have led to strong performance on molecular understanding and generation tasks, yet these gains often come without reliable structural grounding. In particular, existing approaches conflict with the chemistry principle that structure determines function: despite their downstream success, current molecular LLMs perform poorly on basic structure recognition, suggesting that they fail to capture molecular graphs from canonical SMILES. To remedy this, we propose MolBasic, a structure-first framework that strengthens structural comprehension via SMILES–Graph translation. MolBasic is built around a multi-level structure perception benchmark, where bidirectional SMILES–Graph conversion serves as the core task to align sequential and topological representations. On top of this foundation, we employ a progressive learning scheme with a standardized Chain-of-Thought (CoT) to steer models from structure acquisition toward higher-level molecular reasoning. Experiments show that MolBasic substantially improves structural understanding and yields robust gains on downstream tasks, including property prediction and objective optimization, supporting our structure-first paradigm.
\end{abstract}

\section{Introduction}
\label{introduction}
Understanding molecular characteristics and properties through structural analysis, and designing molecules for specific objectives, has long been a central focus of research in biochemistry. This effort is of fundamental importance for advancing drug mechanism studies and the design of novel drugs. With the continuous improvement of Large Language Models (LLMs) in learning and reasoning, more studies are exploring how to leverage the domain knowledge embedded in these models for molecular understanding tasks. Compared to manual analysis, LLMs offer clear advantages in terms of cost and efficiency, making them an ideal tool for this purpose. Therefore, developing an “AI chemist” capable of structural understanding and reasoning of molecular compounds is crucial.

\begin{figure}[t]
\centering
\includegraphics[width=\linewidth]{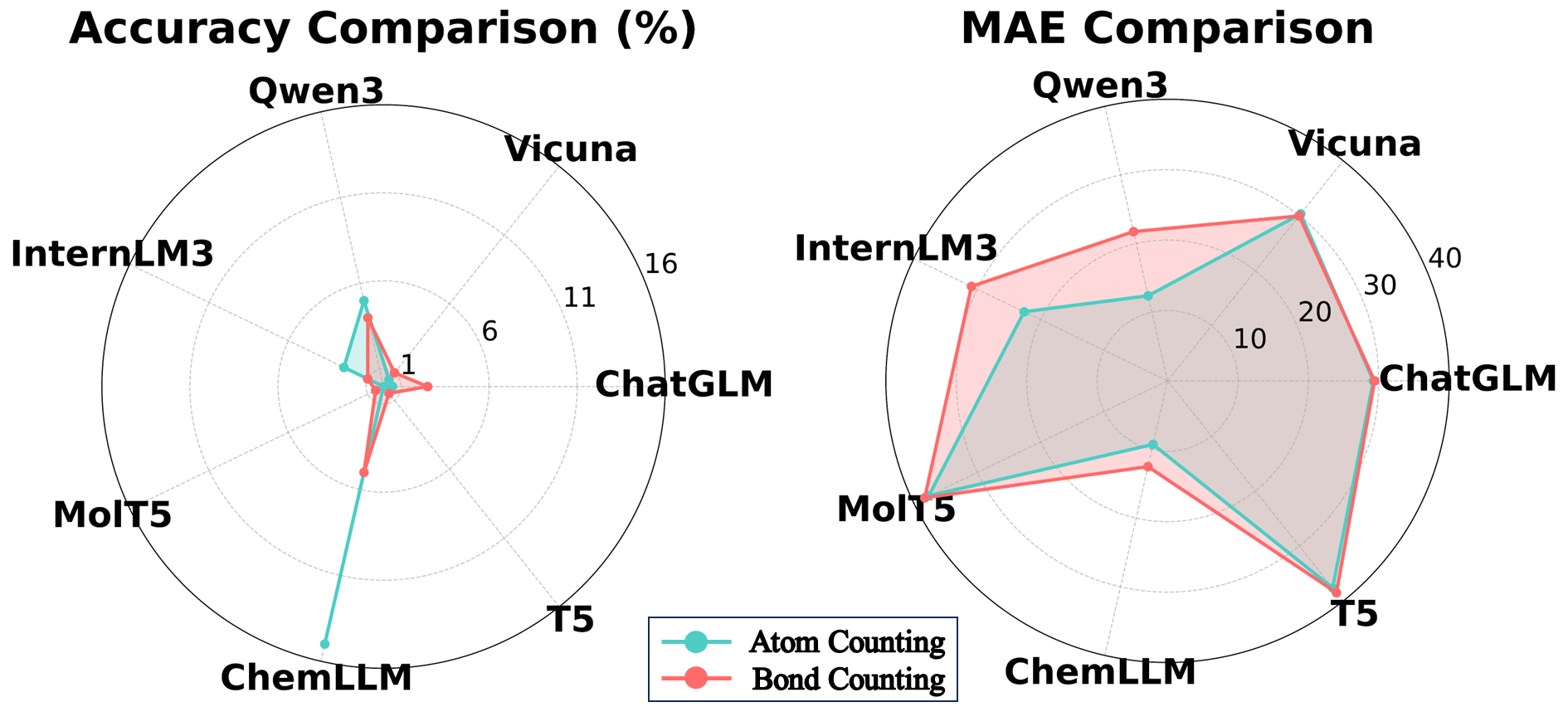}
\caption{Preliminary evaluation of generalist and molecular specialist LLMs on basic structure comprehension tasks (atom counting and bond counting) using a pilot dataset sampled from PubChem \cite{pubchem}. Results reveal that current LLMs fail to capture fundamental graph structure information from SMILES.}
\label{fig:preliminary}
\end{figure}

\begin{figure*}[t]
    \centering
    \includegraphics[width=\linewidth]{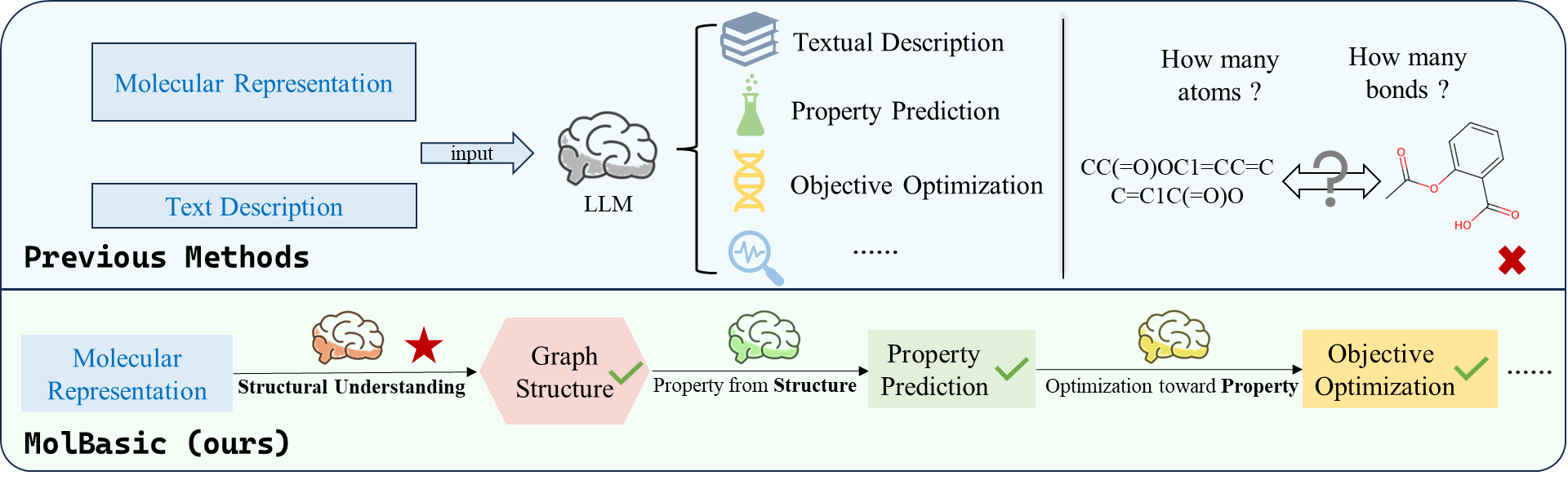}
    \caption{Comparison between the previous paradigm and our MolBasic framework for molecular understanding.}
    \label{fig:paradigm}
\end{figure*}
Generalist LLMs, such as GPT~\cite{gpt3,gpt4}, Qwen~\cite{qwen}, T5~\cite{t5}, InternLM~\cite{internlm}, Vicuna~\cite{vicuna}, and ChatGLM~\cite{chatglm}, acquire knowledge across diverse domains during pre-training and thus possess certain molecular understanding capabilities. However, their knowledge is primarily derived from medical or biological texts related to molecules, which grants them basic knowledge retrieval abilities, rather than genuine structural comprehension. Of course, models with strong reasoning and mathematical capabilities, such as GPT-5~\cite{gpt5}, can achieve molecular understanding through tool invocation and mathematical computation. Molecular specialist LLMs, on the other hand, are the mainstream approach for LLM-based molecular understanding. Existing research predominantly focuses on two directions: (i) designing molecular representations that facilitate model comprehension---for instance, HIGHT~\cite{hight} employs multi-level molecular representations at atom and motif levels to highlight subgraph-level features, while InstructMol~\cite{instructmol} and MoMu~\cite{momu} use graph encoders to embed molecular graphs into input tokens; (ii) designing domain-specific post-training tasks on base models---for instance, MolT5~\cite{molt5} replaces corrupted spans between SMILES and text, Atomas~\cite{atomas} and MoleculeSTM~\cite{moleculestm} align SMILES-text token representations, while GIMLET~\cite{gimlet} and ChemLLM~\cite{chemllm} focus on chemical question answering.

Current molecular LLMs appear to have achieved substantial progress, demonstrating strong performance on common molecular understanding tasks including molecular captioning, property prediction, reaction prediction, and (conditional) molecular generation. However, our preliminary experiments (shown in \cref{fig:preliminary}) reveal that both generalist and specialist models perform surprisingly poorly on fundamental structural perception tasks, such as heavy atom counting and bond counting. Although prior SMILES parsing methods \cite{cleanmol, chemcotbench} attempt to improve chemical understanding by learning local syntactic patterns or substructure-level features, they still fall short of establishing explicit equivalence between sequential SMILES and the complete 2D molecular graph. This indicates that current LLMs fail to capture the graph structure information underlying canonical SMILES \cite{smiles}---information that is crucial for subsequent molecular reasoning and represents the most natural and direct cognition that human chemists have of molecules. These findings suggest that the current understanding paradigm of molecular LLMs contradicts the widely accepted principle in chemistry: \textit{structure determines function}. We argue that molecular LLMs should not learn high-level tasks immediately after simple pre-training on chemical texts, as this approach lacks logical interpretability. Instead, they should first focus on establishing a clear understanding of molecular structure before proceeding to downstream reasoning tasks.

Building on the progress and problems identified above, we propose MolBasic (\textbf{B}asic \textbf{S}tructure \textbf{I}dentification and \textbf{C}omprehension), a back-to-basics framework that grounds molecular understanding in fundamental structure comprehension. With SMILES–Graph mutual conversion as the core task, MolBasic enables LLMs to establish equivalence between sequential and topological molecular representations, building a solid foundation for higher-level reasoning. Our contributions can be summarized as follows:
\begin{itemize}
    \item \textbf{Structure-first Reasoning}: We construct a multi-level Molecular Structure Comprehension benchmark (MSC) comprising eight tasks, with SMILES–Graph mutual conversion as the core task, evaluating LLMs’ fundamental 2D graph perception capabilities for reliable downstream reasoning.
    
    \item \textbf{Progressive Framework}: We propose a staircase reasoning framework that mirrors the cognitive process of human chemists, progressively advancing from structure to property to optimization. For each stage, we design Chain-of-Thought protocols combining powerful LLMs with expert knowledge to enhance structure-based reasoning capabilities.
    
    \item \textbf{Enhanced Reasoning}: Through sufficient experiments across structural property prediction, molecular objective optimization, and few-shot bioactivity prediction tasks, we demonstrate the superior performance and practical applicability of our structure-first reasoning and stepwise framework.
\end{itemize}
\cref{fig:paradigm} illustrates the comparison between our paradigm and previous methods in molecular understanding, emphasizing key distinctions.

\section{Related Work}
\label{related_work}

\subsection{Molecular Understanding}
Early approaches employed task-specific architectures such as GNNs \cite{gcn,gat,graphormer} for property prediction and sequence models \cite{seq2seq,chemformer} for molecular optimization. More recently, molecular LLMs have emerged that generalize across tasks through chemical pretraining \cite{kv-plm,gimlet}, instruction tuning \cite{mol-instructions,chemllm,molrag}, or multimodal integration \cite{chemvlm}. In contrast, our work focuses on post-training LLMs to enhance their structure-aware reasoning abilities, strengthening the internal understanding of molecular structure as a foundation for downstream reasoning, rather than aligning chemical text or augmenting retrieval capabilities.

\subsection{Reasoning Enhancement for LLMs}
Chain-of-Thought (CoT) reasoning has evolved from a prompting technique \cite{wei2022chain,kojima2022large} to a core capability of advanced LLMs. OpenAI o1/o3 \cite{o1} and DeepSeek-R1 \cite{deepseek-r1} incorporate long-chain CoT during training, while GPT-5 \cite{gpt5} further integrates an intelligent router that automatically invokes deep reasoning for complex queries. Progressive learning, drawing from curriculum learning \cite{bengio2009curriculum}, enhances reasoning by organizing training according to task dependencies and difficulty \cite{xu2020curriculum}, proving effective in mathematical problem solving \cite{lightman2023let} and multi-hop reasoning \cite{press2022measuring}. In our work, we combine CoT of strong reasoning models with expert knowledge, proposing the staircase progressive learning framework, starting from basic tasks and following the reasoning process of human chemists.
\definecolor{skyblue}{RGB}{135, 206, 235}
\section{Method}
\label{method}
\begin{figure*}[t]
\centering
\includegraphics[width=\textwidth]{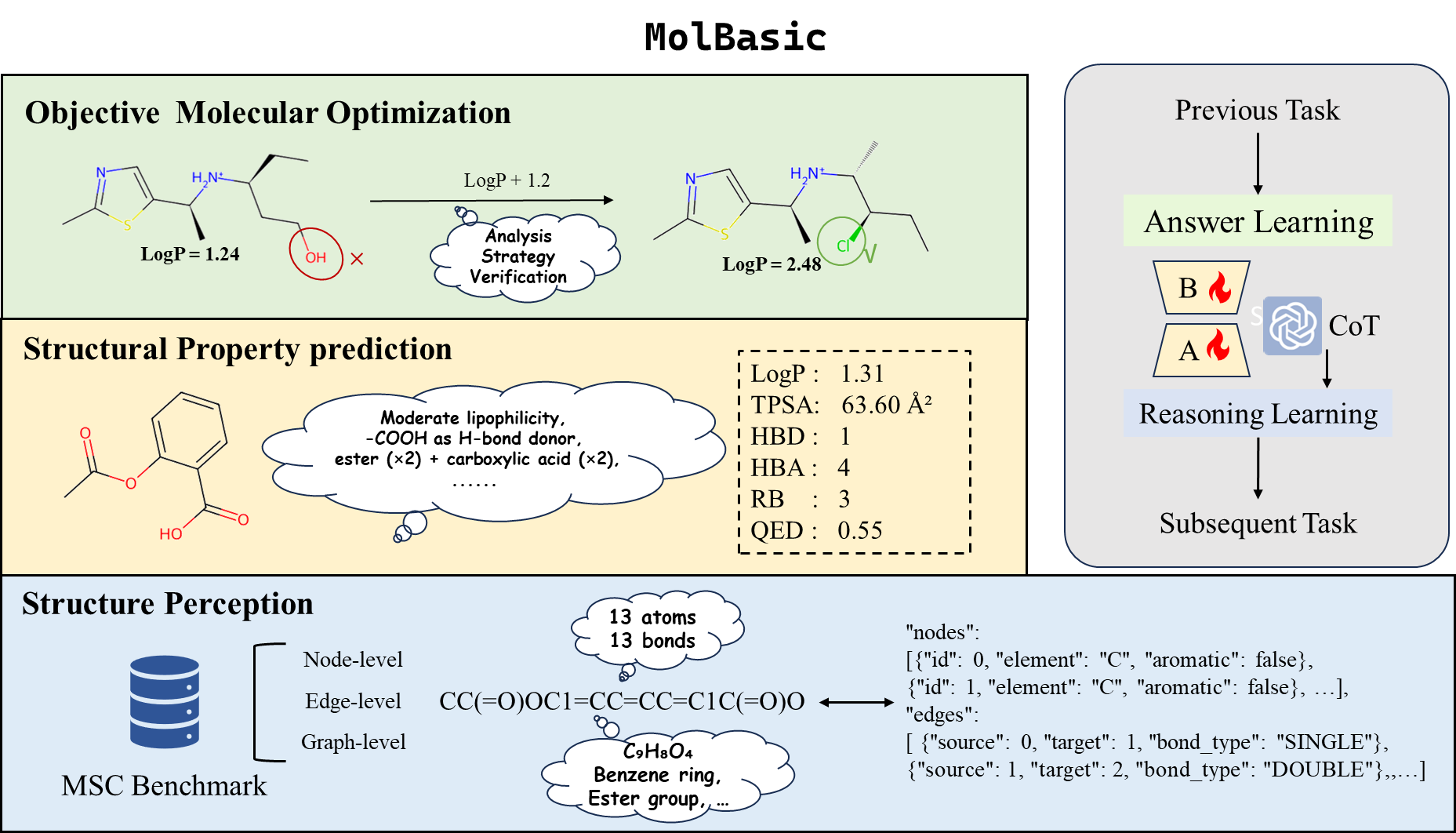}
\caption{Overview of the MolBasic framework. \textbf{Left:} The staircase learning path progresses through three stages: (1) structure comprehension at node, edge, and graph levels using the MSC benchmark; (2) Structural Property Prediction deriving molecular properties from structural features; (3) Objective Molecular Optimization modifying molecules toward desired properties. \textbf{Right:} The two-stage training strategy for each task, where Stage 1 (Answer Learning) establishes correct output anchors, and Stage 2 (Reasoning Learning) incorporates Chain-of-Thought supervision distilled from strong reasoning models.}
\label{fig:pipeline}
\end{figure*}
The complete workflow of MolBasic is illustrated in \cref{fig:pipeline}. In \cref{3.1}, we introduce the structure comprehension benchmark designed to enable LLMs to understand molecular structures. \cref{3.2} presents our staircase learning framework for molecular understanding. \cref{3.3} describes the design of reasoning Chain-of-Thought for molecular understanding tasks. \cref{3.4} details the comprehensive training strategies.
\subsection{Multi-level Structure Comprehension Benchmark}
\label{3.1}
Canonical SMILES serves as the primary input for molecular LLMs due to its text-based format that facilitates tokenization. However, most existing models treat SMILES as plain text sequences, performing global semantic alignment with textual inputs, or at best, matching at the motif or substructure level. For human chemists, the more critical information encoded in SMILES representations is the molecular graph structure. Therefore, to enable LLMs to understand and analyze molecular structures more precisely rather than superficially, models must first learn the equivalence between canonical SMILES and 2D molecular topology, and be able to correctly articulate the details of this graph structure, such as node and edge information, which are fundamental to chemical reasoning.

Based on this motivation, we construct a \textbf{M}ulti-level \textbf{S}tructure \textbf{C}omprehension QA benchmark (\textbf{MSC}) to equip molecular LLMs with the ability to parse SMILES and understand the SMILES-graph equivalence. Our benchmark comprises eight tasks $\mathcal{T}_{struct}=\{T_1, T_2, \cdots, T_8\}$ organized across three levels. At the node level, models are required to identify the number of heavy atoms and the count of specific atom types. At the edge level, we assess whether models understand molecular connectivity by querying the total number of bonds and counts of specific bond types. At the graph level, molecular formula conversion serves as a comprehensive evaluation of node and edge understanding, as models must count each atom type and infer the number of hydrogen atoms based on heavy atom connectivity to satisfy chemical valence rules. Substructure recognition is also essential, as functional groups often serve as the fundamental units of molecular function. Most importantly, we design bidirectional conversion tasks between canonical SMILES and textual graph adjacency lists, enabling LLMs to directly learn the equivalence between molecular representations and 2D graph structures. We process the PubChemSTM dataset \cite{pubchem, moleculestm} and obtain 185,286 valid molecules. Detailed dataset splits are described in Section 4.1.

\subsection{Staircase Learning Framework}
\label{3.2}
To ensure that higher-level molecular reasoning is built upon reliable structural understanding, we organize molecular tasks into a \textbf{staircase learning framework that explicitly models their dependency relationships.}

We view the natural logic of molecular understanding as a DAG-structured reasoning process rooted in structure comprehension. Formally, let $\mathcal{G} = (\mathcal{V}, \mathcal{E})$ denote a directed acyclic graph (DAG) representing the logical relationships among all tasks, where each node $v \in \mathcal{V}$ represents a molecular understanding task, and each directed edge $(v_i, v_j) \in \mathcal{E}$ indicates that task $v_j$ depends on the relevant knowledge and reasoning capabilities acquired from $v_i$. The root node $v_0$ corresponds to structure comprehension, serving as the foundation for all downstream reasoning.

Let $\mathcal{C}$ denote the reasoning capability of the model. For any task $v$ with prerequisite tasks $\text{Pa}(v) = \{u \in \mathcal{V} \mid (u, v) \in \mathcal{E}\}$, the model first acquires capability $\mathcal{C}_{\text{Pa}(v)}$ by learning the prerequisite tasks, then subsequently obtains new capability through learning task $v$ as follows:
\vspace{-1pt}
\begin{equation}
\mathcal{C}_v = f_v(\mathcal{C}_{\text{Pa}(v)}, \mathcal{D}_v)
\end{equation}
\vspace{-1pt}
where $f_v$ denotes the learning process on task $v$, and $\mathcal{D}_v$ is the knowledge data for task $v$. For the root node, $\mathcal{C}_{v_0} = f_{v_0}(\mathcal{C}_0, \mathcal{D}^{\text{CoT}}_{\text{struct}})$ represents the fundamental capability of structure comprehension, where $\mathcal{C}_0$ is the initial capability of the base model prior to fine-tuning.

In this work, we study linear task sequences $\mathcal{P} = (v_0, v_1, ..., v_K)$ where $(v_{k-1}, v_k) \in \mathcal{E}$, meaning that the reasoning knowledge required by $K$ tasks exhibits a linear progressive relationship:
\vspace{-1pt}
\begin{equation}
\mathcal{T}_0 \prec \mathcal{T}_1 \prec \cdots \prec \mathcal{T}_{k-1} \prec \mathcal{T}_k \prec \cdots \prec \mathcal{T}_K
\end{equation}
\vspace{-1pt}
The $k$-th task is learned by an LLM that has integrated the reasoning capabilities from the preceding $k-1$ tasks.

In this paper, we demonstrate the learning and reasoning process of our staircase learning framework through the path $\mathcal{P}: v_{\text{struct}} \rightarrow v_{\text{prop}} \rightarrow v_{\text{opt}}$, which chains together a series of representative molecular understanding tasks:
\begin{equation}
\begin{aligned}
\mathcal{C}_{\text{struct}} &= f_{\text{struct}}(\mathcal{C}_0, \mathcal{D}_{\text{struct}}) \\
\mathcal{C}_{\text{prop}} &= f_{\text{prop}}(\mathcal{C}_{\text{struct}}, \mathcal{D}_{\text{prop}}) \\
\mathcal{C}_{\text{opt}} &= f_{\text{opt}}(\mathcal{C}_{\text{prop}}, \mathcal{D}_{\text{opt}})
\end{aligned}
\end{equation}
This path aligns with the cognitive process of human chemists and reflects the fundamental chemistry principle that \textit{structure determines function}.

\subsection{Chain-of-Thought Construction}
\label{3.3}
To improve the accuracy and interpretability of molecular reasoning, we introduce explicit Chain-of-Thought supervision that \textbf{constrains intermediate structural analysis during training.} In traditional molecular understanding task training, models typically learn only the mapping from input molecule $S$ to final prediction $a$, i.e., optimizing the probability $P_\theta(a \mid S)$. Although models can directly learn input-to-output mappings, in complex tasks, training without explicit reasoning supervision leads to reduced model accuracy and lack of interpretability. Therefore, we enhance the knowledge QA dataset by introducing Chain-of-Thought (CoT) distilled from strong reasoning models. Specifically, we distill reasoning trajectories from GPT-5 \cite{gpt5} to construct the CoT supervision.

By incorporating the reasoning trajectory $z$ as a supervisory signal, we extend the training objective to jointly model the conditional distribution over molecular representation, reasoning process, and results:
\begin{equation}
P_\theta(a, z \mid S, q) = P_\theta(z \mid S, q) \cdot P_\theta(a \mid S, z, q)
\end{equation}
\vspace{-1pt}
Compared to supervision through final answers alone, this approach effectively constrains the hypothesis space and reduces ambiguity in the reasoning process, thereby improving reasoning accuracy and stability. Joint supervision ensures that the loss function considers not only the final output but also the prediction error of the reasoning process.

We construct CoTs by distilling reasoning trajectories from strong reasoning models and applying manual review to standardize reasoning steps and remove erroneous trajectories. This ensures reliable, reproducible reasoning. We highlight two key aspects of the resulting CoT design:

(i) Deterministic Intermediate Results for Regression Tasks. For structure comprehension and property prediction, we design CoTs with verifiable intermediate results. For instance, atom counting explicitly parses SMILES into atoms and records counts before aggregation; property prediction first identifies functional groups and structural features, then derives values from these intermediates. This makes each step deterministic and checkable against ground truth.

(ii) Explicit Analysis and Modification Explanation for Optimization Tasks. For molecular optimization, the CoT follows an ``analysis--strategy--verification'' pipeline: it analyzes the original structure and properties, identifies groups affecting the target property and proposes modifications, and verifies the optimized molecule achieves the desired improvement. This encourages learning transferable modification principles rather than memorizing specific edits.

After constructing the molecular reasoning CoT as described above, we merge it into the answer and use $\mathcal{D}^{\text{CoT}}_v$ as shown in \cref{eq:cot_data} as the knowledge dataset for training:
\vspace{-1pt}
\begin{equation}
\hat{a} = (z, a), \quad \mathcal{D}^{\text{CoT}}_v = \{(S^{j}, (q^{j}, \hat{a}^{j}))\}_{j=1}^{N_v}
\label{eq:cot_data}
\end{equation}
\vspace{-1pt}
By augmenting training labels with Chain-of-Thought explanations from strong reasoning models, we transform molecular reasoning from a black-box mapping into a structured prediction problem with explicit supervision.

\begin{table*}[!ht]
\centering
\caption{Performance comparison on multi-level structure comprehension benchmark. For MAE, lower is better ($\downarrow$); for ACC, higher is better ($\uparrow$). \colorbox{teal!30}{\textbf{Best}} and \colorbox{skyblue!50}{second best} results are highlighted.}
\label{tab:structure_perception}
\setlength{\tabcolsep}{8pt}
\renewcommand{\arraystretch}{1}
\resizebox{\textwidth}{!}{
\begin{tabular}{llccccccccc}
\toprule
Level & Tasks & Metrics & Vicuna & ChatGLM & T5 & Qwen3-8B & InternLM3 & MolT5 & ChemLLM & MolBasic \\
\midrule
\multirow{4}{*}{Node} 
& \multirow{2}{*}{Heavy Atom Counting} 
& MAE $\downarrow$ & 28.804 & 28.228 & 33.535 & 11.893 & 19.754 & 35.04 & \colorbox{skyblue!50}{6.696} & \colorbox{teal!30}{\textbf{1.223}} \\
& & ACC $\uparrow$ & 1.21\% & 0.73\% & 0.22\% & 6.09\% & 4.75\% & 0.4\% & \colorbox{skyblue!50}{14.81\%} & \colorbox{teal!30}{\textbf{42.44\%}} \\
\cmidrule{2-11}
& \multirow{2}{*}{Element-specific Counting} 
& MAE $\downarrow$ & 8.511 & 9.127 & 9.882 & \colorbox{skyblue!50}{2.982} & 3.902 & 9.833 & 3.763 & \colorbox{teal!30}{\textbf{0.339}} \\
& & ACC $\uparrow$ & 11.49\% & 12.22\% & 0.13\% & \colorbox{skyblue!50}{51.86\%} & 40.98\% & 2.5\% & 42.1\% & \colorbox{teal!30}{\textbf{73.01\%}} \\
\midrule
\multirow{4}{*}{Edge} 
& \multirow{2}{*}{Total Bond Counting} 
& MAE $\downarrow$ & 29.85 & 26.506 & 35.19 & 18.352 & 30.265 & 35.755 & \colorbox{skyblue!50}{10.581} & \colorbox{teal!30}{\textbf{1.158}} \\
& & ACC $\uparrow$ & 1.08\% & 1.86\% & 0.09\% & 3.07\% & 1.12\% & 0.4\% & \colorbox{skyblue!50}{3.41\%} & \colorbox{teal!30}{\textbf{49\%}} \\
\cmidrule{2-11}
& \multirow{2}{*}{Specific Bond Counting} 
& MAE $\downarrow$ & 9.8 & 10.557 & 10.227 & \colorbox{skyblue!50}{7.674} & 8.105 & 9.287 & 36.67 & \colorbox{teal!30}{\textbf{0.383}} \\
& & ACC $\uparrow$ & 22.58\% & 14.12\% & 2.81\% & \colorbox{skyblue!50}{45.81\%} & 26.34\% & 3.3\% & 5.76\% & \colorbox{teal!30}{\textbf{82.9\%}} \\
\midrule
\multirow{4}{*}{Graph} 
& Formula Convert & ACC $\uparrow$ & 0\% & 0\% & 0\% & 0\% & 0\% & 0\% & 0\% & \colorbox{teal!30}{\textbf{44.04\%}} \\
\cmidrule{2-11}
& Substructure Recognition & ACC $\uparrow$ & 48.75\% & 46.03\% & 0.13\% & \colorbox{skyblue!50}{72.67\%} & 66.23\% & 19.8\% & 10.71\% & \colorbox{teal!30}{\textbf{98.14\%}} \\
\cmidrule{2-11}
& SMILES to Graph & ACC $\uparrow$ & 0\% & 0\% & 0\% & 0\% & 0\% & 0\% & \colorbox{skyblue!50}{0.04\%} & \colorbox{teal!30}{\textbf{94.65\%}} \\
\cmidrule{2-11}
& Graph to SMILES & ACC $\uparrow$ & 0\% & 0\% & 0\% & 0\% & 0\% & 0\% & 0\% & \colorbox{teal!30}{\textbf{85.66\%}} \\
\bottomrule
\end{tabular}
}
\end{table*}

\subsection{Training Strategy}
\label{3.4}
We present the multi-stage training of MolBasic based on LoRA fine-tuning. First, following the staircase learning framework proposed in \cref{3.2}, we sequentially train each task along the task path $\mathcal{P}: v_{\text{struct}} \rightarrow v_{\text{prop}} \rightarrow v_{\text{opt}}$ on a base model (Qwen3-8B \cite{qwen3} in this work).

For training each individual task, we follow the easy-to-hard curriculum learning principle and divide the training into two stages. Stage 1 focuses on learning the answer $a$ only, dedicated to domain knowledge acquisition. Stage 2 learns the complete response containing CoT, dedicated to refining reasoning capabilities. This two-stage training first establishes correct answer anchors to constrain the reasoning process towards the correct direction, then forms standardized and formalized reasoning processes that further enhance accuracy and interpretability through explicit and verifiable intermediate steps.

Both stages employ LoRA-based next token prediction with cross-entropy loss as shown in \cref{eq:loss_stage1} and \eqref{eq:loss_stage2}. The training algorithm of MolBasic is presented in Appendix~\ref{algorithm}.
\vspace{-3pt}
\begin{align}
\mathcal{L}_{\text{stage1}} &= -\sum_{t=1}^{|a|} \log P_\theta(a_t \mid S, q, a_{<t}) \label{eq:loss_stage1}\\
\mathcal{L}_{\text{stage2}} &= -\sum_{t=1}^{|\hat{a}|} \log P_\theta(\hat{a}_t \mid S, q, \hat{a}_{<t}) \label{eq:loss_stage2}
\end{align}
\vspace{-6pt}

\definecolor{skyblue}{RGB}{135, 206, 235}

\section{Experiments}
\label{experiments}
In this section, we present experimental results of MolBasic. We first describe training and evaluation details on our Multi-level structure comprehension (MSC) benchmark in \cref{4.1}. Then in \cref{4.2}, we demonstrate how the acquired structural perception capabilities transfer to downstream molecular understanding tasks through progressive reasoning, including structural property prediction (\cref{4.2.1}), objective molecular optimization (\cref{4.2.2}), and few-shot transfer to bioactivity prediction tasks (\cref{4.2.3}). Training parameters and training/inference costs for each stage are listed in \cref{tab:training_config} (\cref{appendix:training}).
\subsection{Structure Comprehension}
\label{4.1}
\textbf{Datasets and Baselines:} As introduced in \cref{3.1}, our MSC benchmark uses the PubChemSTM dataset reported by MoleculeSTM \cite{moleculestm}, following the same preprocessing pipeline with raw data sourced from PubChem \cite{pubchem}. In Stage 1, weights are assigned based on task importance, with higher weights given to the two SMILES-graph conversion tasks. In Stage 2, weights are adjusted according to Stage 1 performance, increasing weights for underperforming tasks. Detailed training set construction strategies for both stages are provided in Appendix \ref{appendix:training}. For efficient and unbiased evaluation, we sample 2,316 instances from the full test set to construct a benchmark that preserves the original distribution. We compare against mainstream general-purpose and molecular LLMs with similar parameter scales, including Qwen3-8B \cite{qwen3}, InternLM3-8B \cite{internlm}, T5-Large \cite{t5}, Vicuna-7B \cite{vicuna}, ChatGLM3-6B \cite{chatglm}, MolT5-Large \cite{molt5}, and ChemLLM-7B \cite{chemllm}.

\begin{table*}[t]
\centering
\caption{Results on structural property prediction. \colorbox{teal!30}{\textbf{Best}} and \colorbox{skyblue!50}{second best} results are highlighted among LLM-based methods. ``--'' indicates unreasonable or invalid outputs, implying failure in property prediction.}
\label{tab:property_prediction}
\resizebox{\textwidth}{!}{
\begin{tabular}{l|cc|cc|cc|cc|cc|cc|cc}
\toprule
\multirow{2}{*}{\textbf{Method}} & \multicolumn{2}{c|}{\textbf{MW}} & \multicolumn{2}{c|}{\textbf{LogP}} & \multicolumn{2}{c|}{\textbf{TPSA}} & \multicolumn{2}{c|}{\textbf{HBD}} & \multicolumn{2}{c|}{\textbf{HBA}} & \multicolumn{2}{c|}{\textbf{RB}} & \multicolumn{2}{c}{\textbf{QED}} \\
& Pear.$\uparrow$ & MAE$\downarrow$ & Pear.$\uparrow$ & MAE$\downarrow$ & Pear.$\uparrow$ & MAE$\downarrow$ & Pear.$\uparrow$ & MAE$\downarrow$ & Pear.$\uparrow$ & MAE$\downarrow$ & Pear.$\uparrow$ & MAE$\downarrow$ & Pear.$\uparrow$ & MAE$\downarrow$ \\
\midrule
Chemception (CNN) & 0.87 & 44.75 & 0.62 & 1.23 & 0.84 & 14.45 & 0.81 & 0.58 & 0.84 & 0.93 & 0.73 & 1.91 & -0.004 & 0.40 \\
Chemprop (GNN) & 0.99 & 1.63 & 0.98 & 0.16 & 0.99 & 0.91 & 0.99 & 0.03 & 0.99 & 0.11 & 0.99 & 0.15 & 0.89 & 0.07 \\
\midrule
Qwen-VL-7B & 0.78 & 58.07 & 0.14 & 1.48 & 0.19 & 82.83 & -0.02 & 1.42 & 0.05 & 4.91 & 0.19 & 6.87 & -- & -- \\
InternVL-v1.5-20B & 0.59 & 83.60 & 0.04 & 2.37 & 0.29 & 28.73 & 0.03 & 2.24 & 0.22 & 2.41 & 0.04 & 4.82 & 0.003 & 0.40 \\
LLaVA-v1.5-7B & 0.36 & 115.70 & -0.003 & 1.61 & 0.01 & 99.14 & 0.004 & 3.74 & 0.04 & 2.93 & 0.03 & 3.85 & -- & -- \\
\midrule
ChemVLM-8B & 0.84 & 56.94 & 0.38 & 1.68 & 0.26 & 53.66 & 0.49 & 1.35 & 0.32 & 4.58 & 0.10 & 5.56 & -0.003 & 0.37 \\
ChemMLLM-7B & 0.97 & 16.17 & 0.92 & 0.52 & 0.97 & 6.06 & 0.94 & 0.13 & 0.94 & 0.44 & 0.94 & 1.62 & 0.91 & 0.06 \\
ChemMLLM-34B & \colorbox{skyblue!50}{0.98} & \colorbox{teal!30}{\textbf{11.57}} & \colorbox{skyblue!50}{0.93} & \colorbox{skyblue!50}{0.43} & \colorbox{skyblue!50}{0.98} & \colorbox{skyblue!50}{3.54} & \colorbox{skyblue!50}{0.96} & \colorbox{skyblue!50}{0.11} & \colorbox{skyblue!50}{0.96} & \colorbox{skyblue!50}{0.26} & \colorbox{skyblue!50}{0.97} & \colorbox{skyblue!50}{0.75} & \colorbox{skyblue!50}{0.93} & \colorbox{teal!30}{\textbf{0.05}} \\
\midrule
MolBasic & \colorbox{teal!30}{\textbf{0.996}} & \colorbox{skyblue!50}{13.98} & \colorbox{teal!30}{\textbf{0.97}} & \colorbox{teal!30}{\textbf{0.28}} & \colorbox{teal!30}{\textbf{0.99}} & \colorbox{teal!30}{\textbf{2.71}} & \colorbox{teal!30}{\textbf{0.99}} & \colorbox{teal!30}{\textbf{0.05}} & \colorbox{teal!30}{\textbf{0.99}} & \colorbox{teal!30}{\textbf{0.16}} & \colorbox{teal!30}{\textbf{0.98}} & \colorbox{teal!30}{\textbf{0.36}} & \colorbox{teal!30}{\textbf{0.94}} & \colorbox{skyblue!50}{0.06} \\
\bottomrule
\end{tabular}
}
\end{table*}

\textbf{Results:} \cref{tab:structure_perception} presents the performance of various models on multi-level structure comprehension tasks. MolBasic achieves substantial improvements across all tasks, with the most striking gains on bidirectional SMILES-graph conversion---a fundamental yet previously overlooked capability where all existing LLMs universally fail with near-zero accuracy. Since the graph adjacency list format is uncommon, we provide a simple example in the instruction as an output template (applied fairly to all models). Conversion accuracy is verified using RDKit~\cite{rdkit} by checking exact molecular match against the ground truth.

\subsection{Downstream Molecular Understanding}
\label{4.2}
After establishing structure comprehension, we evaluate MolBasic on representative high-level molecular tasks. Beyond the progressive relationship in \cref{3.2}, task selection prioritizes structure-based reasoning over memorization or knowledge retrieval. We focus on property prediction and molecular optimization, which require structural analysis rather than text generation tasks.
\subsubsection{Structural Property Prediction}
\label{4.2.1}
Structural properties refer to molecular attributes determined by the molecular structure. They can be predicted by analyzing key atoms, bonds, and functional groups, including MW, LogP, TPSA, HBD, HBA, RB, and QED. The detailed definitions of these properties are provided in Appendix~\ref{appendix:properties}.

We fine-tune and evaluate our model on 100K molecules sampled from the PubChem dataset. Baseline methods include task-specific deep learning models based on traditional architectures (Chemception \cite{chemception}, ChemProp \cite{chemprop}), as well as representative (multi-modal) large language models, including Qwen-VL \cite{qwenvl}, InternVL \cite{internvl}, LLaVA \cite{llava}, and ChemVLM \cite{chemvlm}. \cref{tab:property_prediction} reports the prediction errors and Pearson correlation coefficients between the predicted values and ground-truth labels for structural property prediction tasks.

As shown in \cref{tab:property_prediction}, MolBasic consistently achieves the best performance across almost all properties, significantly outperforming other LLM-based methods, while achieving performance comparable to task-specific models. These results demonstrate that once an LLM is equipped with strong structure comprehension ability, it can predict structure-related molecular properties more accurately and reliably.

\subsubsection{Objective Molecular Optimization}
\label{4.2.2}
Based on the model's understanding of structural properties learned in the previous stage, we equip MolBasic with the ability to optimize molecular structures toward a desired objective, such as improving drug-likeness or enhancing solubility. Specifically, we take the task of increasing LogP as a representative example, and use 175K molecules from the TDC dataset~\cite{tdc}.
\begin{table}[!ht]
\centering
\caption{Performance comparison on molecular optimization. 
For all 3 metrics, higher is better ($\uparrow$). 
\colorbox{teal!30}{\textbf{Best}} and \colorbox{skyblue!50}{second best} results are highlighted among LLM-based methods.}
\label{tab:logp_optimization}
\setlength{\tabcolsep}{4pt}
\renewcommand{\arraystretch}{1}
\resizebox{\columnwidth}{!}{
\begin{tabular}{lccc}
\toprule
Model & Increased LogP ($\uparrow$) & Diversity ($\uparrow$) & Validity ($\uparrow$) \\
\midrule
Seq2Seq & 1.95 & 0.79 & 80.5\% \\
ChemFormer & 3.03 & 0.85 & 100\% \\
\midrule
Qwen-VL-7B & 1.50 & \colorbox{skyblue!50}{0.95} & 4.0\% \\
InternVL-v1.5-20B & 0.77 & 0.90 & 48.0\% \\
LLaVA-v1.5-7B & 1.72 & \colorbox{teal!30}{\textbf{0.96}} & 37.5\% \\
GPT-4o & \colorbox{skyblue!50}{1.97} & 0.86 & \colorbox{teal!30}{99.0\%} \\
ChemVLM-8B & 0.67 & 0.87 & 92.5\% \\
\midrule
MolBasic & \colorbox{teal!30}{\textbf{2.27}} & 0.86 & \colorbox{skyblue!50}{\textbf{98.7\%}} \\
\bottomrule
\end{tabular}
}
\end{table}

\begin{table*}[t]
\centering
\caption{Ablation studies. We examine four ablation settings against the full MolBasic pipeline to isolate the contribution of each design choice. ``Skipped'' indicates that the corresponding task stage is bypassed under that setting and thus not evaluated.}
\label{tab:ablation}
\resizebox{\textwidth}{!}{
\begin{tabular}{l|ccccc|ccccccc|ccc}
\toprule
\multirow{2}{*}{\textbf{Setting}}
& \multicolumn{5}{c|}{\textbf{Structure Comprehension}}
& \multicolumn{7}{c|}{\textbf{Property Prediction (MAE$\downarrow$)}}
& \multicolumn{3}{c}{\textbf{Molecular Optimization}} \\
& \makecell{Atom\\MAE$\downarrow$} & \makecell{Atom\\ACC$\uparrow$} & \makecell{Bond\\MAE$\downarrow$} & \makecell{Bond\\ACC$\uparrow$} & \makecell{Form.\\ACC$\uparrow$} 
& MW & LogP & TPSA & HBD & HBA & RB & QED
& $\Delta$LogP$\uparrow$ & Div.$\uparrow$ & Val.$\uparrow$ \\
\midrule
MolBasic & \textbf{1.22} & \textbf{42.4} & \textbf{1.16} & \textbf{49.0} & \textbf{44.0} & \textbf{13.98} & \textbf{0.28} & \textbf{2.71} & \textbf{0.05} & \textbf{0.16} & \textbf{0.36} & \textbf{0.06} & \textbf{2.27} & 0.86 & 98.7\% \\
\midrule
\multicolumn{16}{l}{\textit{(a) Is structure comprehension a necessary foundation?}} \\
\midrule
w/o Struct  & \multicolumn{5}{c|}{\textit{Skipped}} & 42.29 & 0.64 & 14.42 & 0.62 & 1.19 & 0.64 & 0.10 & 1.80 & 0.86 & 97.6\% \\
Direct Opt  & \multicolumn{5}{c|}{\textit{Skipped}} & \multicolumn{7}{c|}{\textit{Skipped}} & 1.41 & 0.85 & 94.2\% \\
\midrule
\multicolumn{16}{l}{\textit{(b) Does the staircase ordering matter, beyond total data volume?}} \\
\midrule
Mixed SFT   & \multicolumn{5}{c|}{\textit{Skipped}} & 122.87 & 1.85 & 35.49 & 0.89 & 1.87 & 2.82 & 0.25 & 0.98 & \textbf{0.87} & 95.3\% \\
\midrule
\multicolumn{16}{l}{\textit{(c) What role does CoT supervision play?}} \\
\midrule
w/o CoT     & 1.51 & 16.4 & 1.61 & 11.2 & 22.1 & 35.67 & 0.53 & 11.12 & 0.59 & 0.71 & 0.53 & 0.08 & 2.05 & 0.85 & \textbf{99.5\%} \\
\bottomrule
\end{tabular}
}
\end{table*}

For baseline methods, we include traditional sequence-based models, Seq2Seq \cite{seq2seq} and ChemFormer \cite{chemformer}, as well as representative LLM-based approaches, incorporating GPT-4o \cite{gpt4o} for comparison. \cref{tab:logp_optimization} reports the performance of our MolBasic on the molecular optimization task with LogP increase as the objective, compared to other methods.

As shown in \cref{tab:logp_optimization}, MolBasic achieves the best LogP improvement among LLM-based models while maintaining high validity and diversity. As the highest-level task in our stepwise reasoning framework, we present example responses with explanation for the optimization task in Appendix \ref{appendix:interpretability}, demonstrating the interpretability of our approach.

\subsection{Ablation Study}
\label{sec:ablation}

In this section, we validate the key design choices of MolBasic through ablation experiments organized around three questions: (i)~whether structure comprehension serves as a necessary foundation for downstream tasks, (ii)~whether the staircase learning order itself drives performance gains beyond simply increasing training data, and (iii)~what role Chain-of-Thought supervision plays in the framework. All results are consolidated in Table~\ref{tab:ablation}.

\textbf{(a) Is structure comprehension a necessary foundation?}
A central claim of MolBasic is that structural perception acquired in the first stage provides essential capabilities for all subsequent tasks. To test this, we construct two settings that progressively strip away prerequisite stages: \textbf{w/o Struct}, which skips structure comprehension and directly trains the base model on property prediction followed by optimization, and \textbf{Direct Opt}, which trains the base model on optimization alone without any prerequisite. As shown in Table~\ref{tab:ablation}(a), performance degrades monotonically as more stages are removed. The degradation from MolBasic to w/o Struct demonstrates that even when the model receives the same downstream fine-tuning, the absence of prior structural understanding leads to clearly inferior property prediction. The further decline from w/o Struct to Direct Opt confirms that each stage in the staircase contributes foundational capabilities that cannot be compensated by task-specific training alone.

\textbf{(b) Does the staircase ordering matter?} 
Although the results in Table~\ref{tab:ablation}(a) already show that skipping prerequisite reasoning stages in the staircase process negatively affects performance, we design the \textbf{Mixed SFT} setting to disentangle this effect from the total training data volume. We merge the training sets of both downstream tasks (property prediction and molecular optimization) and apply the same two-stage curriculum learning strategy (answer learning followed by reasoning learning), ensuring that the total number of supervised signals is matched. The only difference is the removal of progressive task ordering. As Table~\ref{tab:ablation}(b) shows, Mixed SFT leads to reduced performance across metrics, confirming that staircase ordering is essential—when reasoning processes from different task levels are learned simultaneously, they interfere with each other, preventing the model from establishing the structured knowledge hierarchy.

\textbf{(c) What role does CoT supervision play?}
Finally, we examine the contribution of Chain-of-Thought reasoning supervision by comparing full MolBasic with a \textbf{w/o CoT} variant that preserves progressive training order but removes Stage 2 reasoning learning at all stages, retaining only answer-level supervision from Stage 1. As Table~\ref{tab:ablation}(c) shows, removing CoT supervision leads to consistent performance drops across structure comprehension and downstream tasks. Structure comprehension tasks, such as atom counting and formula conversion, require multi-step reasoning, and CoT provides explicit guidance for these intermediate steps. Consequently, the absence of CoT reduces the model’s reasoning capabilities, since it lacks explicit supervision on intermediate reasoning steps. This results in lower accuracy on property prediction tasks and decreased effectiveness in molecular optimization, highlighting the importance of CoT in guiding multi-step molecular reasoning. We note that GPT-5, serving as the source of CoT, was also evaluated directly on these tasks (see Appendix); results show that MolBasic becomes competitive with GPT-5 as the reasoning process unfolds.

Additional evaluations and analyses, covering external benchmarks, GPT-5 comparisons, and statistical analysis, are provided in Appendices~\ref{app:additional_experiments}--\ref{appendix:additional_analysis}.
\section{Conclusion}
\label{conclusion}
In this work, we revisit a fundamental but overlooked problem in molecular LLMs: the lack of basic structure comprehension. We demonstrate that without accurately perceiving molecular graphs, higher-level reasoning tasks are inherently unreliable. To address this, we equip LLMs with explicit structural comprehension ability, and then introduce Chain-of-Thought supervision and a progressive learning framework to transfer this capability to downstream tasks. Extensive experiments confirm that restoring this basic structural understanding is the key to achieving accurate, interpretable, and generalizable molecular reasoning.

\section*{Limitations}
One limitation of the current MolBasic framework lies in its staircase reasoning design. In this work, molecular reasoning is modeled as a chain-like process that progresses from structure comprehension to property prediction and then to molecular optimization. As downstream tasks become farther from the initial structure comprehension stage, the direct influence of structural perception may gradually weaken. This is reflected in our results, where the improvement on molecular optimization is less pronounced than that on property prediction.
A promising future direction is to design a more flexible reasoning process that more closely resembles how human chemists analyze molecules, allowing LLMs to explicitly use molecular structure as the foundation for each molecular understanding task. However, this is non-trivial: simply adding structure comprehension data to every downstream task is unlikely to be sufficient, as our ablation study shows that directly mixing tasks can introduce interference. Designing a principled structure-grounded reasoning framework for diverse molecular tasks remains an important direction for future work.

\section*{Ethical Considerations}
This work focuses on AI-assisted molecular understanding and beneficial molecular optimization. In our experiments, optimization objectives are limited to standard drug-relevant properties, such as improving LogP, QED, and synthetic accessibility, which are commonly used to assess molecular quality and developability. We do not optimize toward toxic, harmful, or unsafe molecular objectives. During CoT post-training, we also observed that the base model retained safety-aware behavior for potentially hazardous molecules, sometimes refusing to provide related information. This suggests that the safety alignment of the base model can serve as an additional safeguard against unsafe molecular generation. For future real-world applications, generated molecules should still be subject to standard safety screening, toxicity filtering, and human expert review before deployment. 

\bibliography{latex/reference}
\newpage
\appendix

\section*{Appendices}
\section{Algorithm}
\label{algorithm}
Algorithm~\ref{alg:MolBasic} summarizes the complete training procedure of \textsc{MolBasic}, which follows the staircase learning framework and adopts a two-stage optimization strategy for each task. Starting from a base language model $\mathcal{C}_0$, the algorithm progressively equips the model with increasingly advanced molecular reasoning capabilities along a predefined task path $\mathcal{P} = (v_{\text{struct}}, v_{\text{prop}}, v_{\text{opt}})$.
\begin{algorithm}[htbp]
\caption{MolBasic Training Algorithm}
\label{alg:MolBasic}
\begin{algorithmic}[1]
\REQUIRE Base model $\mathcal{C}_0$, task path $\mathcal{P} = (v_{\text{struct}}, v_{\text{prop}}, v_{\text{opt}})$, datasets $\{\mathcal{D}_v, \mathcal{D}^{\text{CoT}}_v\}_{v \in \mathcal{P}}$
\ENSURE Fine-tuned model with capability $\mathcal{C}_{\text{opt}}$

\STATE Initialize LoRA parameters $\theta$
\STATE $\mathcal{C}_{\text{prev}} \leftarrow \mathcal{C}_0$

\FOR{each task $v_k$ in $\mathcal{P}$}
    \STATE \textbf{// Stage 1: Answer Learning}
    \FOR{each $(S, (q, a))$ in $\mathcal{D}_{v_k}$}
        \STATE Compute $\mathcal{L}_{\text{stage1}} = -\sum_{t=1}^{|a|} \log P_\theta(a_t \mid S, q, a_{<t})$
        \STATE Update $\theta$ via gradient descent
    \ENDFOR
    \STATE \textbf{// Stage 2: Reasoning Learning}
    \FOR{each $(S, (q, \hat{a}))$ in $\mathcal{D}^{\text{CoT}}_{v_k}$}
        \STATE Compute $\mathcal{L}_{\text{stage2}} = -\sum_{t=1}^{|\hat{a}|} \log P_\theta(\hat{a}_t \mid S, q, \hat{a}_{<t})$
        \STATE Update $\theta$ via gradient descent
    \ENDFOR
    \STATE $\mathcal{C}_{v_k} \leftarrow f_{v_k}(\mathcal{C}_{\text{prev}}, \mathcal{D}^{\text{CoT}}_{v_k})$
    \STATE $\mathcal{C}_{\text{prev}} \leftarrow \mathcal{C}_{v_k}$
\ENDFOR

\STATE \textbf{return} Model with capability $\mathcal{C}_{\text{opt}}$
\end{algorithmic}
\end{algorithm}

For each task $v_k \in \mathcal{P}$, training is performed in two consecutive stages. \textbf{Stage~1 (Answer Learning)} focuses on acquiring task-specific domain knowledge by supervising only the final answers. In this stage, the model is optimized to correctly generate task outputs given the molecular input $S$ and query $q$, establishing reliable output anchors and preventing early-stage reasoning drift.

\textbf{Stage~2 (Reasoning Learning)} further refines the model by incorporating Chain-of-Thought (CoT) supervision. Instead of predicting only the final answer, the model is trained to generate the complete reasoning trajectory $\hat{a} = (z, a)$, where $z$ represents intermediate reasoning steps. This stage explicitly constrains the reasoning process, reduces ambiguity in intermediate representations, and improves both accuracy and interpretability.

After completing both stages for task $v_k$, the updated model capability $\mathcal{C}_{v_k}$ is obtained and passed as initialization to the next task in the staircase. In this manner, structural comprehension learned at earlier stages is preserved and reused when learning higher-level tasks such as structural property prediction and molecular optimization. After all tasks in $\mathcal{P}$ are completed, the algorithm returns a model with capability $\mathcal{C}_{\text{opt}}$, which integrates structure-aware reasoning across all stages.

\section{Data Preparation and Training Details}
\label{appendix:training}
We prepare the training data from the PubChemSTM dataset~\cite{moleculestm}, covering all eight structure comprehension tasks (see Table~\ref{tab:tasks}): atom counting, total bond counting, specific bond counting, element counting, substructure recognition, SMILES to graph conversion, graph to SMILES conversion, and formula generation.
For data cleaning, we first remove duplicate samples across all tasks. For the two SMILES-Graph conversion tasks, we further filter out samples whose graph representation exceeds 16K characters, which empirically corresponds to exceeding the 2048-token context limit under our tokenizer. After preprocessing, the dataset contains 1,181,816 valid samples expanded across eight tasks, corresponding to 185,286 unique molecules. The expanded samples are split into training, validation, and test sets.

We summarize the dataset sources and sizes used for all molecular understanding tasks in Table~\ref{tab:dataset_summary}. 

\begin{table*}[!ht]
\centering
\caption{Dataset statistics for different molecular understanding tasks. Values in parentheses indicate the expanded sample counts after constructing MSC tasks.}
\label{tab:dataset_summary}
\begin{tabular}{lccccc}
\toprule
Task & Dataset & Total & Training & Valid & Test \\
\midrule
\makecell{Structure comprehension\\(expand by MSC)}
& PubChemSTM
& \makecell{185,286\\(1,181,816)}
& \makecell{129,699\\(737,828)}
& \makecell{37,059\\(296,012)}
& 18,528 \\
\midrule
Property prediction & PubChem & 100,000 & 95,000 & -- & 5,000 \\
Molecular optimization & TDC & 175,173 & 157,673 & -- & 17,500 \\
\bottomrule
\end{tabular}
\end{table*}

\begin{table}[!ht]
\centering
\caption{Multi-level structure comprehension tasks.}
\label{tab:tasks}
\resizebox{\columnwidth}{!}{
\begin{tabular}{lll}
\toprule
\textbf{Level} & \textbf{Task} & \textbf{Task Type} \\ 
\midrule
\multirow{2}{*}{Node} & Atom Counting & Numerical Regression\\
                      & Element Counting & Numerical Regression\\
\midrule
\multirow{2}{*}{Edge} & Total Bond Counting & Numerical Regression\\
                      & Specific Bond Counting & Numerical Regression\\
\midrule
\multirow{4}{*}{Graph} & Formula Conversion & Sequence Generation \\
                       & Substructure Recognition & Binary Classification\\
                       & Graph to SMILES & Sequence Generation\\
                       & SMILES to Graph & Sequence Generation\\
\bottomrule
\end{tabular}
}
\end{table}

Stage 1 (Answer Learning): To emphasize the importance of SMILES-Graph mutual conversion as the core task, we apply a weighted sampling strategy. The two conversion tasks are assigned a weight of 2.0, while all other tasks have a weight of 1.0, resulting in approximately doubled sampling frequency for the conversion tasks.

Stage 2 (Reasoning Learning): We construct a smaller-scale dataset by sampling 1/10 from the Stage 1 validation set. Task weights are adjusted based on Stage 1 performance to prioritize underperforming tasks: atom counting (3.0), total bond counting (3.0), formula generation (3.0), SMILES to graph (3.0), graph to SMILES (3.0), element counting (1.5), specific bond counting (1.5), and substructure recognition (0.5). Importantly, Chain-of-Thought reasoning is only applied to the three tasks that showed suboptimal performance in Stage 1—atom counting, total bond counting, and formula generation—while other tasks retain the answer-only training format from Stage 1. The Stage 1 validation set is used exclusively for Stage 2 training and is not involved in model evaluation, ensuring no data leakage.

We provide the detailed training configurations for each stage in \cref{tab:training_config}. Our progressive training follows a sequential pipeline: Multi-level Structure Comprehension (MSC) → Structural Property Prediction (Prop) → Molecular Optimization (Optim), where each task consists of two stages—Stage 1 (Answer Learning) and Stage 2 (Reasoning Learning). The LoRA adapter from each stage serves as the initialization for the subsequent stage, enabling knowledge accumulation throughout the training process. All stages share the same LoRA architecture (rank=64, alpha=128) and learning rate (1e-4) with cosine scheduling, while batch sizes and training epochs are adjusted based on dataset sizes and task complexity.

We also report training and inference efficiency. The MSC-S1 stage, which processes the largest dataset, requires approximately 6 days on 4$\times$48 GB RTX 4090 GPUs. All subsequent stages are significantly more efficient, each completing within a few hours on 2$\times$80 GB A800 GPUs. For inference, all inference times are measured with batch size 1, except Optim-S2. For Optim-S2, full CoT generation takes about 84s per sample with \texttt{max\_new\_tokens=1024} at batch size 1; the reported 7.089s is obtained by batched inference. This demonstrates that our CoT reasoning approach incurs minimal additional computational cost in practical deployment scenarios.

\textbf{Artifact use: }We use publicly available datasets, models, and tools for research purposes and follow their original terms of use. 

\begin{table*}[t]
\centering
\caption{Training configurations across different stages. MSC: Multi-level Structure Comprehension, Prop: Structural Property Prediction, Optim: Molecular Optimization. S1: Answer Learning, S2: Reasoning Learning.}
\label{tab:training_config}
\setlength{\tabcolsep}{4pt}
\footnotesize
\begin{tabular}{lcccccc}
\toprule
\textbf{Parameter} & \textbf{MSC-S1} & \textbf{MSC-S2} & \textbf{Prop-S1} & \textbf{Prop-S2} & \textbf{Optim-S1} & \textbf{Optim-S2} \\
\midrule
Base Model & Qwen3-8B & +MSC-S1 & +MSC-S2 & +Prop-S1 & +Prop-S2 & +Optim-S1 \\
LoRA Rank & \multicolumn{6}{c}{64} \\
LoRA Alpha & \multicolumn{6}{c}{128} \\
LoRA Dropout & \multicolumn{6}{c}{0.05} \\
\midrule
Learning Rate & \multicolumn{6}{c}{1e-4} \\
Effective Batch Size & 32 & 16 & 32 & 128 & 128 & 128 \\
Epochs & 1 & 5 & 2 & 5 & 2 & 2 \\
LR Scheduler & \multicolumn{6}{c}{cosine} \\
Warmup Ratio & \multicolumn{6}{c}{0.1} \\
\midrule
Max Seq Length & \multicolumn{6}{c}{2048} \\
DeepSpeed & - & ZeRO-2 & - & - & - & - \\
\midrule
Steps & 5188 & 1850 & 1336 & 1665 & 1108 & 1108 \\
FLOPs & 2.32e19 & 2.86e18 & 3.39e18 & 2.06e19 & 3.20e18 & 1.00e19 \\
Samples/sec & 3.27 & 4.31 & 5.36 & 4.75 & 5.50 & 5.04 \\
Steps/sec & 0.026 & 0.067 & 0.042 & 0.018 & 0.021 & 0.020 \\
\midrule
Infer. Time (s) & 0.301 & 0.311 & 5.139 & 4.977 & 6.863 & 7.089\\
\bottomrule
\end{tabular}
\end{table*}

\section{Definitions of Molecular Properties}
\label{appendix:properties}
In this section, we provide concise definitions of the structural properties used in our experiments, all of which are directly derived from molecular structure.
\begin{itemize}
    \item \textbf{Molecular Weight (MW):} The sum of the atomic weights of all atoms in a molecule, reflecting its overall molecular size.
    \item \textbf{Octanol--Water Partition Coefficient (LogP):} The logarithm of the ratio of a compound’s concentration in octanol to that in water, measuring its hydrophobicity.
    \item \textbf{Topological Polar Surface Area (TPSA):} The surface area contributed by polar atoms (typically oxygen and nitrogen and their attached hydrogens), indicating a molecule’s polarity and permeability.
    \item \textbf{Hydrogen Bond Donors (HBD):} The number of functional groups in a molecule capable of donating a hydrogen atom to form a hydrogen bond.
    \item \textbf{Hydrogen Bond Acceptors (HBA):} The number of atoms in a molecule capable of accepting a hydrogen bond through lone pairs.
    \item \textbf{Number of Rotatable Bonds (RB):} The count of non-ring, single bonds between heavy atoms that allow free rotation, characterizing molecular flexibility.
    \item \textbf{Quantitative Estimate of Drug-likeness (QED):} A composite score that quantifies drug-likeness by integrating multiple physicochemical properties using a desirability function.
\end{itemize}

\section{Evaluation on External Benchmarks}
\label{app:additional_experiments}

To verify that the gains of MolBasic are not limited to our proposed Molecular Structure Comprehension benchmark, we further evaluate it on two external molecular understanding benchmarks for fair comparison. First, the CleanMol benchmark ~\cite{cleanmol} focuses on SMILES parsing tasks, including functional-group recognition, ring counting, chain-length prediction, and canonical SMILES generation. As shown in Table~\ref{tab:smiles_parsing_performance}, MolBasic achieves the best accuracy on most tasks, substantially outperforming both few-shot general LLMs and most SFT baselines, demonstrating strong general structural parsing ability.

Second, ChemCoTBench~\cite{chemcotbench} provides a broader evaluation of molecular understanding, including functional-group counting, ring counting, Murcko scaffold extraction, ring-system scaffold extraction, and SMILES equivalence judgment. As shown in Table~\ref{tab:chemcotbench_molecular_understanding}, MolBasic achieves the best performance on most metrics and remains competitive on ring counting. These results further confirm that MolBasic learns transferable structural understanding beyond our proposed benchmark.
\begin{table*}[!ht]
\centering
\caption{SMILES parsing performance on CleanMol benchmark. All metrics are reported as accuracy.}
\label{tab:smiles_parsing_performance}
\small
\setlength{\tabcolsep}{5pt}
\begin{tabular}{llcccc}
\toprule
Task type & Model & FG & Ring & Chain & Canonical \\
\midrule
\multirow{3}{*}{5-shot}
& Deepseek-V3-chat & 0.8912 & 0.6266 & 0.2976 & 0.1484 \\
& GPT-4o           & 0.8750 & 0.5955 & 0.2857 & 0.1078 \\
& Galactica-6.7B   & 0.5000 & 0.0732 & 0.1511 & 0.0000 \\
\midrule
\multirow{5}{*}{SFT}
& Llama3.1-8B (Single) & 0.9414 & 0.8612 & 0.9859 & 0.9356 \\
& Llama3.1-8B (Multi)  & 0.9891 & 0.8707 & 0.9851 & \textbf{0.9463} \\
& Qwen2.5-7B (Single)  & 0.9891 & 0.8674 & 0.9907 & 0.7593 \\
& Qwen2.5-7B (Multi)   & 0.9901 & 0.8750 & 0.9902 & 0.9262 \\
& MolBasic & \textbf{0.9956} & \textbf{0.9774} & \textbf{0.9946} & 0.8836 \\
\bottomrule
\end{tabular}
\end{table*}

\begin{table*}[!ht]
\centering
\caption{ChemCoTBench molecular understanding evaluation compared with molecular LLMs. The evaluation uses MAE for functional-group and ring counting, Tanimoto similarity for Murcko scaffold extraction, and Accuracy for ring-system, SMILES equivalence.}
\label{tab:chemcotbench_molecular_understanding}
\small
\setlength{\tabcolsep}{6pt}
\begin{tabular}{lccccc}
\toprule
\multirow{2}{*}{Models}
& \multicolumn{2}{c}{Func-Group}
& \multicolumn{2}{c}{Scaffold}
& SMILES \\
\cmidrule(lr){2-3}
\cmidrule(lr){4-5}
\cmidrule(lr){6-6}
& FG$\downarrow$ & Ring$\downarrow$
& Murcko$\uparrow$ & Ring-sys$\uparrow$
& Eq.$\uparrow$ \\
\midrule
Ether0
& Failed & \textbf{0.35}
& Failed & Failed
& 0.63 \\
BioMedGPT-7B
& 1.6 & 2.43
& 0.18 & 0.53
& 0.39 \\
BioMistral-7B
& 1.0 & 1.85
& 0.04 & 0.33
& 0.50 \\
MolBasic
& \textbf{0.73} & 0.65
& \textbf{0.43} & \textbf{0.65}
& \textbf{0.67} \\
\bottomrule
\end{tabular}
\end{table*}

\section{Few-shot in Bioactivity Prediction Tasks}
\label{4.2.3}
\begin{table}[!ht]
\centering
\caption{Few-shot results on bioactivity prediction tasks. \colorbox{teal!30}{\textbf{Best}} and \colorbox{skyblue!50}{second best} results are highlighted among few-shot methods.}
\label{tab:bioactivity}
\setlength{\tabcolsep}{3pt}
\renewcommand{\arraystretch}{1.1}
\footnotesize
\begin{tabular}{llcccc}
\toprule
Shot & Method & BACE & HIV & BBBP & Tox21 \\
\midrule
\rowcolor{gray!20} \multicolumn{6}{l}{\textit{Few-shot Methods}} \\
\multirow{2}{*}{0-shot} & MolRAG & 0.504 & 0.514 & 0.546 & 0.534\\
& MolBasic & 0.500 & 0.499 & 0.514 & 0.563 \\
\cmidrule{1-6}
\multirow{2}{*}{1-shot} & MolRAG & 0.594 & 0.543 & 0.546 & 0.545\\
& MolBasic & 0.619 & \colorbox{teal!30}{\textbf{0.664}} & 0.550 & 0.561 \\
\cmidrule{1-6}
\multirow{2}{*}{2-shot} & MolRAG & 0.615 & 0.568 & 0.564 & 0.55 \\
& MolBasic & \colorbox{skyblue!50}{0.649} & 0.654 & \colorbox{skyblue!50}{0.611} & \colorbox{teal!30}{\textbf{0.579}} \\
\cmidrule{1-6}
\multirow{2}{*}{4-shot} & MolRAG & 0.626 & 0.595 & 0.572 & 0.566 \\
& MolBasic & \colorbox{teal!30}{\textbf{0.661}} & \colorbox{skyblue!50}{0.659} & \colorbox{teal!30}{\textbf{0.633}} & \colorbox{skyblue!50}{0.573} \\
\midrule
\rowcolor{gray!20} \multicolumn{6}{l}{\textit{Pre-training Methods}} \\
& GIMLET & 0.696 & 0.662 & 0.594 & 0.612\\
& KVPLM & 0.513 & 0.612 & 0.602 & 0.492\\
& MoMu & 0.666 & 0.503 & 0.498 & 0.576\\
& Galactica-1.3B & 0.565 & 0.339 & 0.539 & 0.495\\
\midrule
\rowcolor{gray!20} \multicolumn{6}{l}{\textit{Graph-based Networks}} \\
& GCN & 0.736 & 0.757 & 0.649 & 0.749\\
& GAT & 0.697 & 0.729 & 0.665 & 0.754\\
& GIN & 0.701 & 0.753 & 0.658 & 0.74\\
& Graphormer & 0.776 & 0.745 & 0.702 & 0.759\\
\bottomrule
\end{tabular}
\vspace{-10pt}
\end{table}
Beyond fine-tuning on downstream tasks, we conduct few-shot experiments on tasks logically related but not seen during training, to demonstrate the transferability of our framework to novel yet relevant tasks after acquiring fundamental capabilities. We select bioactivity property prediction, closely related to molecular structural properties, and compare with MolRAG, the current state-of-the-art few-shot method. As shown in \cref{tab:bioactivity}, MolBasic achieves better performance on most properties and approaches the performance of pre-training methods. We provide descriptions of the baseline categorization used in \cref{tab:bioactivity} in Appendix \ref{appendix:fewshot}.

\begin{table*}[!ht]
\centering
\caption{Comparison of MolBasic and GPT-5 on structure comprehension tasks. Metrics are accuracy ($\uparrow$).}
\label{tab:appendix_gpt5_structure}
\resizebox{\textwidth}{!}{
\begin{tabular}{lcccccccc}
\toprule
Model & Atom & Elem. & Total Bond & Spec. Bond & Formula & Substruct. & S$\rightarrow$G & G$\rightarrow$S \\
\midrule
MolBasic & 42.4\% & 73.01\% & 49\% & 82.9\% & 44.0\% & 98.1\% & 94.7\% & 85.7\% \\
GPT-5 & 95\% & 97.5\% & 71.5\% & 75.5\% & 80.7\% & 85.5\% & 0\% & 1.38\% \\
\bottomrule
\end{tabular}
}
\end{table*}

\begin{table*}[!ht]
\centering
\caption{Comparison of MolBasic and GPT-5 on property prediction tasks. Metrics are MAE ($\downarrow$).}
\label{tab:appendix_gpt5_property}
\begin{tabular}{lccccccc}
\toprule
Model & MW & LogP & TPSA & HBD & HBA & RB & QED \\
\midrule
MolBasic & 35.67 & 0.53 & 11.12 & 0.59 & 0.71 & 0.53 & 0.08 \\
GPT-5 & 8.96 & 0.94 & 12.95 & 0.05 & 0.27 & 1.60 & 1.13 \\
\bottomrule
\end{tabular}
\end{table*}

\begin{table}[!ht]
\centering
\caption{Comparison of MolBasic and GPT-5 on molecular optimization. Higher is better for all metrics.}
\label{tab:appendix_gpt5_optimization}
\begin{tabular}{lccc}
\toprule
Model & $\Delta$LogP & Diversity & Validity \\
\midrule
MolBasic & 2.27 & 0.86 & 98.7\% \\
GPT-5 & 1.89 & 0.87 & 95.5\% \\
\bottomrule
\end{tabular}
\end{table}

\section{Few-shot Baseline Settings}
\label{appendix:fewshot}
In this section, we describe the baseline methods used for few-shot bioactivity prediction and clarify the rationale behind their categorization. We group baseline methods into three categories according to their training paradigm and supervision regime: \emph{few-shot methods}, \emph{pre-training-based methods}, and \emph{graph-based networks}.

\textbf{Few-shot Methods.}  
Few-shot methods are evaluated under the same data-scarce setting as MolBasic, where models are provided with only a limited number of labeled examples per task at inference time. Specifically, we include MolRAG as a representative few-shot baseline, which enhances large language models with retrieval over molecular databases. Both MolRAG and MolBasic are evaluated under identical $k$-shot settings ($k \in \{0,1,2,4\}$), ensuring a fair comparison in terms of supervision level and data availability.

\textbf{Pre-training-based Methods.}  
Pre-training-based methods leverage large-scale labeled or unlabeled molecular datasets to acquire task-relevant knowledge before evaluation. Models such as GIMLET, KVPLM, MoMu, and Galactica-1.3B fall into this category, as they rely on extensive pre-training on molecular or molecule--text corpora. While these methods are not few-shot by design, we include them as reference points to contextualize the performance of few-shot approaches against models with substantially more task-specific prior knowledge.

\textbf{Graph-based Networks.}  
Graph-based networks, including GCN, GAT, GIN, and Graphormer, represent task-specific models trained with full supervision on molecular graphs. These methods serve as upper-bound references, as they have direct access to explicit molecular graph structures and are optimized specifically for bioactivity prediction tasks.

Overall, this categorization highlights the trade-off between supervision strength and model generality, and situates MolBasic within the challenging few-shot regime, where strong structural reasoning must be achieved with minimal labeled data.

\section{GPT-5 Evaluation on Molecular Tasks}
\label{appendix:gpt5_results}

To further contextualize the contribution of Chain-of-Thought supervision, we evaluate GPT-5 directly on structure comprehension, property prediction, and molecular optimization tasks. Tables~\ref{tab:appendix_gpt5_structure}--\ref{tab:appendix_gpt5_optimization}
summarize the direct evaluation of GPT-5 and its comparison with MolBasic.

These results show that GPT-5 provides strong reasoning signals, particularly for numerical and additive properties, validating its use as a source of Chain-of-Thought supervision. MolBasic leverages these distilled signals through progressive, stage-wise learning and CoT supervision. As a result, MolBasic progressively internalizes structural reasoning: it first acquires foundational structure comprehension, then transfers these capabilities to property prediction and molecular optimization tasks. This training pipeline enables MolBasic to achieve competitive performance relative to GPT-5 itself, especially for structure-dependent properties such as LogP and QED, while maintaining better performance in molecular optimization.

\section{Additional Analysis}
\label{appendix:additional_analysis}
\subsection{Error Analysis}
\label{appendix:error_analysis}

\begin{figure}[!ht]
    \centering
    \includegraphics[width=0.5\textwidth]{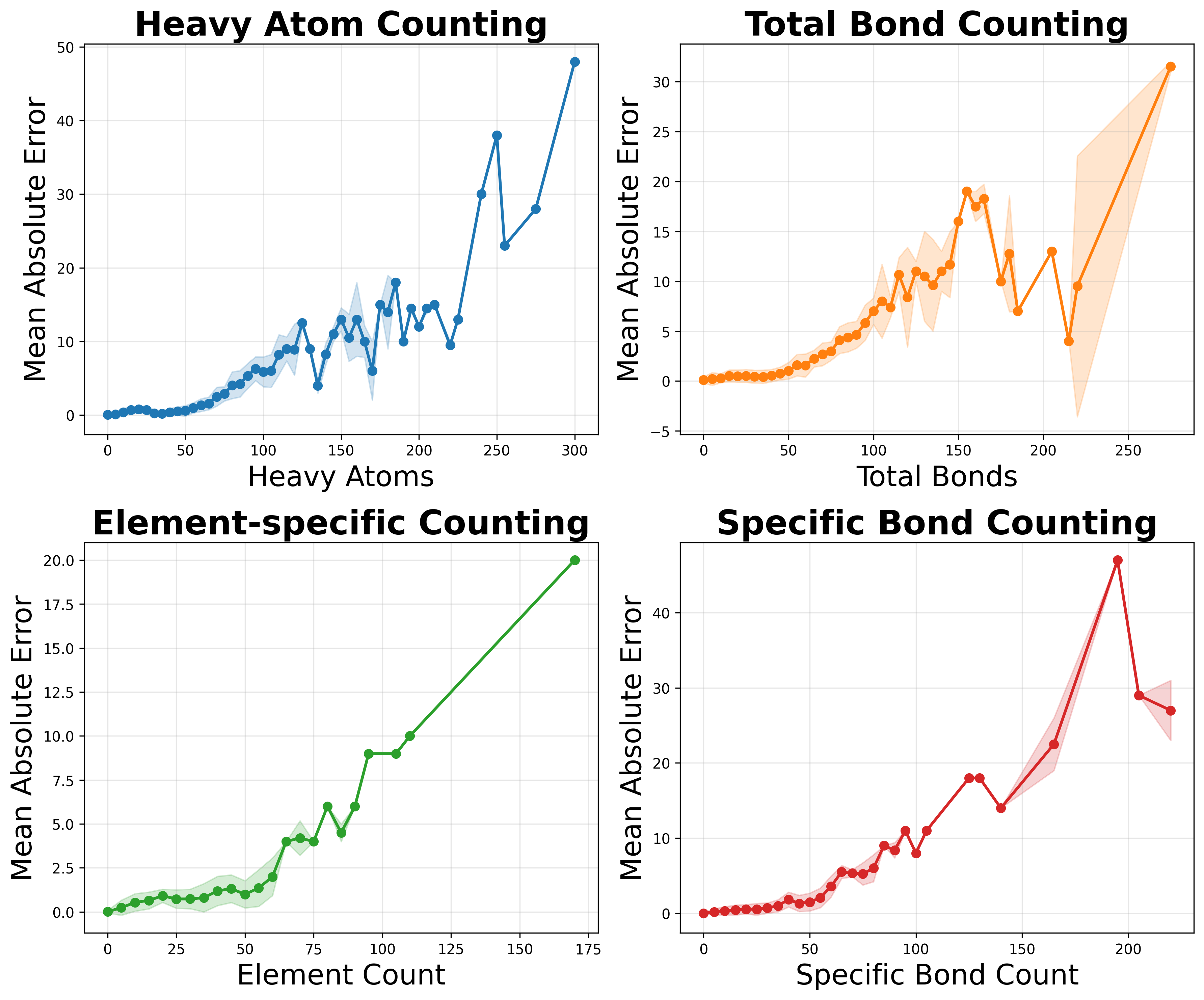}
    \caption{Error distribution across four regression tasks in the multi-level structure comprehension benchmark. Each panel shows the mean absolute error (MAE) as a function of ground-truth values for the corresponding task: (a) Heavy Atom Counting, (b) Total Bond Counting, (c) Element-specific Counting, and (d) Specific Bond Counting. The shaded regions represent the standard deviation of errors within each bin.}
    \label{fig:error_distribution}
\end{figure}

We analyze the sources of prediction errors in numerical regression tasks. As shown in \cref{fig:error_distribution}, prediction errors tend to increase as the ground-truth values become larger, especially for atom and bond counting tasks. This trend mainly arises from two factors. First, LLMs are generally less sensitive to precise numerical values, making exact counting difficult. Second, larger molecules usually correspond to longer and more structurally complex SMILES sequences, containing more rings, branches, and functional groups, which further increases parsing difficulty. These factors lead to prediction errors that scale with molecular size.

\begin{figure*}[!ht]
    \centering
    \includegraphics[width=0.95\textwidth]{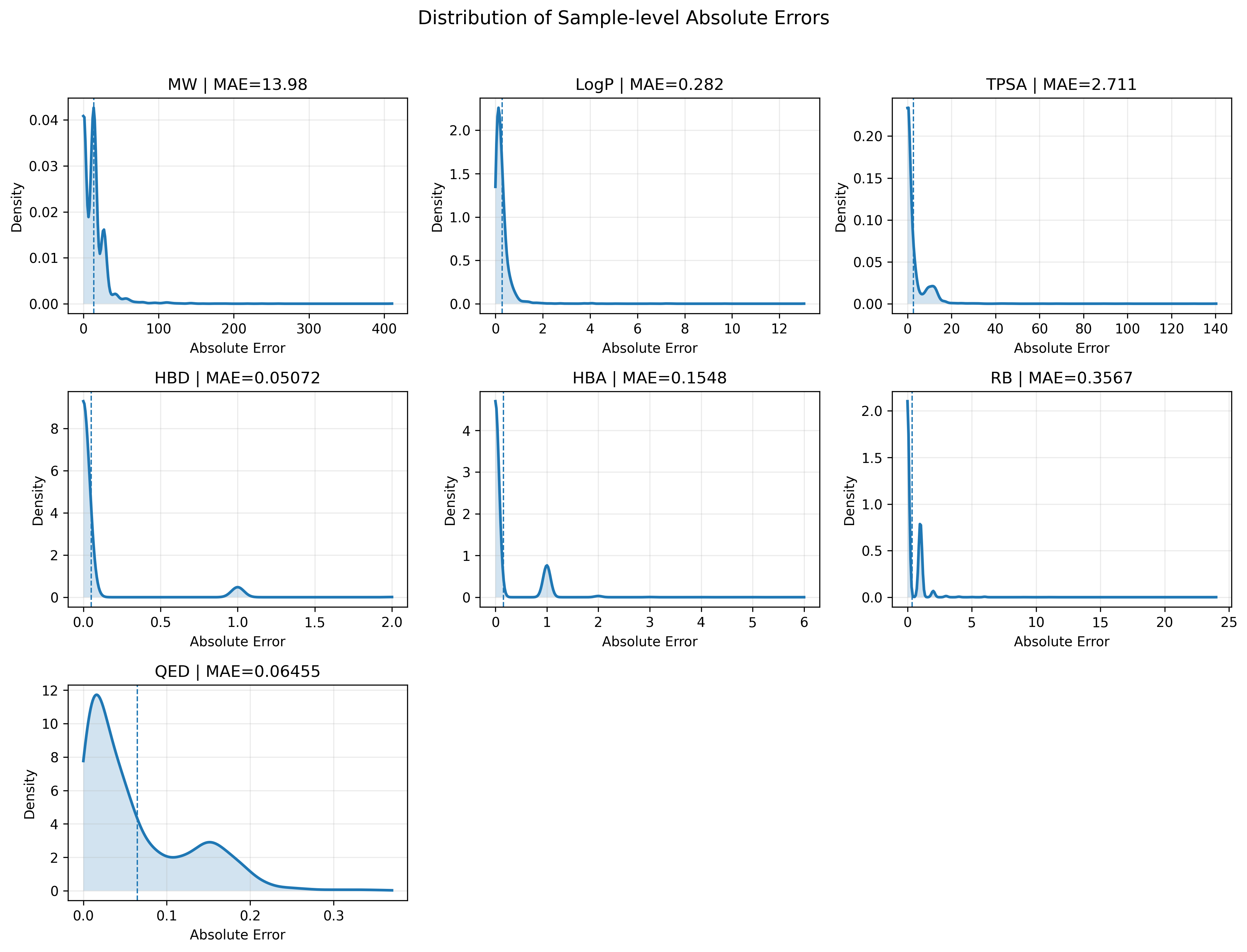}
    \caption{Distribution of sample-level absolute errors for seven molecular properties. Each subplot corresponds to one property. The solid curve shows the density distribution of absolute errors, and the shaded area indicates the filled density region. The vertical dashed line marks the mean absolute error (MAE) of the corresponding property.}
    \label{fig:property_error_distribution}
\end{figure*}

\begin{figure*}[!ht]
    \centering
    \includegraphics[width=0.95\textwidth]{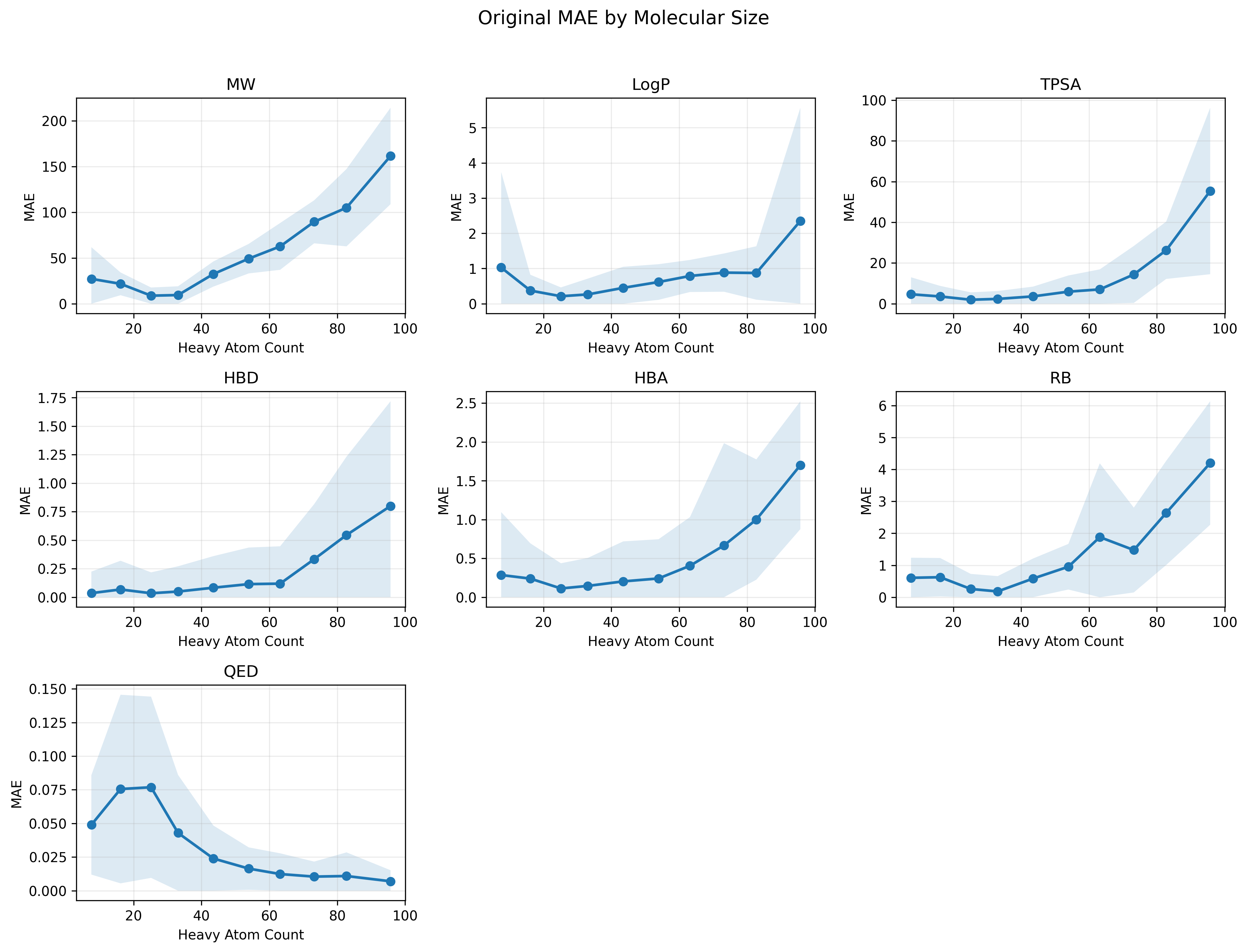}
    \caption{Property prediction errors with respect to molecular size. Each subplot shows the mean absolute error (MAE) of one property across bins of heavy atom counts. The line with markers represents the average MAE within each bin, and the shaded region denotes the standard deviation of errors in that bin.}
    \label{fig:property_error_size}
\end{figure*}

We further analyze MolBasic's prediction errors on molecular property prediction tasks. As shown in \cref{fig:property_error_distribution}, most absolute errors are concentrated near zero, indicating generally accurate predictions, with only a small number of extreme predicted values forming long-tail distributions. \cref{fig:property_error_size} further shows that MolBasic remains stable on molecules of normal size across most properties, while performance degradation mainly appears for larger molecules, where longer SMILES strings and more complex structures make structural parsing and property reasoning more difficult.

\subsection{Multi-objective Molecular Optimization}
\label{appendix:multi_objective_optimization}

In \cref{4.2.2}, we use LogP improvement as the representative objective for molecular optimization. To examine whether optimizing a single property compromises other drug-relevant properties, we further evaluate the changes in QED and synthetic accessibility (SA) under this single-objective setting. As shown in Table~\ref{tab:multi_objective_optimization}, although the model is only instructed to improve LogP, QED decreases only slightly, while SA improves substantially. This indicates that MolBasic, after acquiring molecular structure and property knowledge, can optimize molecules toward the target objective without severely damaging other important chemical properties.

We further conduct a multi-objective fine-tuning experiment, where the model is explicitly trained to increase LogP and QED while decreasing SA. The results show that all three objectives are improved simultaneously: LogP remains substantially increased, QED improves from 0.74 to 0.79 with the QED $\geq 0.8$ ratio increasing from 36.3\% to 57.3\%, and SA decreases from 3.64 to 2.60 with the SA $< 3$ ratio increasing from 23.5\% to 73.8\%.

\begin{table*}[!ht]
\centering
\caption{Multi-objective molecular optimization results. Higher $\Delta$LogP and QED indicate better performance, while lower SA indicates better synthetic accessibility. QED $\geq 0.8$ denotes the percentage of generated molecules with high drug-likeness, and SA $< 3$ denotes the percentage of easy-to-synthesize molecules.}
\label{tab:multi_objective_optimization}
\begin{tabular}{lccccc}
\toprule
Setting & $\Delta$LogP $\uparrow$ & QED $\uparrow$ & QED $\geq 0.8$ $\uparrow$ & SA $\downarrow$ & SA $< 3$ $\uparrow$ \\
\midrule
Single-obj. (LogP only) & +2.27 & 0.72$\rightarrow$0.71 & 39.3\%$\rightarrow$37.6\% & 3.49$\rightarrow$2.65 & 32.3\%$\rightarrow$73.9\% \\
Multi-obj. (LogP+QED+SA) & +2.10 & 0.74$\rightarrow$0.79 & 36.3\%$\rightarrow$57.3\% & 3.64$\rightarrow$2.60 & 23.5\%$\rightarrow$73.8\% \\
\bottomrule
\end{tabular}
\end{table*}

\subsection{Out-of-distribution Analysis}
\label{appendix:ood_analysis}

We conduct Murcko scaffold overlap analysis on both downstream tasks to examine whether MolBasic relies on scaffold memorization. For property prediction, 55.3\% of test scaffolds are unseen during training; for molecular optimization, 24.1\% of test samples are OOD. Table~\ref{tab:ood_analysis} compares the performance on the full test set and scaffold-OOD subsets.

\begin{table*}[!ht]
\centering
\caption{Performance on the full test set and scaffold-OOD subsets. Property prediction metrics are MAE ($\downarrow$), and molecular optimization is measured by $\Delta$LogP ($\uparrow$).}
\label{tab:ood_analysis}
\begin{tabular}{lcccccccc}
\toprule
Subset & MW & LogP & TPSA & HBD & HBA & RB & QED & $\Delta$LogP \\
\midrule
All & 13.98 & 0.28 & 2.71 & 0.05 & 0.16 & 0.36 & 0.06 & 2.27 \\
OOD & 15.02 & 0.30 & 3.07 & 0.06 & 0.18 & 0.33 & 0.06 & 2.11 \\
\bottomrule
\end{tabular}
\end{table*}

The results show that MolBasic maintains stable performance on scaffold-OOD samples. Compared with the full test set, property prediction errors increase only mildly on OOD molecules, and the LogP improvement in molecular optimization decreases slightly from 2.27 to 2.11. Importantly, neither downstream task exhibits performance collapse under scaffold shift, suggesting that MolBasic learns transferable structure-property reasoning rather than merely memorizing training scaffolds.

\section{Future Work}
\label{future_work}
Regarding capability enhancement, further improvements are needed in LLMs' ability to comprehensively comprehend fundamental molecular structures. As our current results indicate, the model still requires strengthening in complex regression tasks and interpretable graph structure conversion. Future work will focus on developing more sophisticated training strategies to better address these challenging aspects of structural perception.

Regarding scope expansion, our next direction aims to approximate the reasoning process of an "AI chemist" by integrating LLMs' structural understanding with efficient retrieval of chemical literature. This would establish a structure-reasoning-driven, knowledge-assisted paradigm for molecular understanding. We believe this combination of structural perception and domain knowledge retrieval will further advance molecular LLMs toward more comprehensive and practical chemical reasoning capabilities.

\section{Chain-of-Thought Templates}
\label{appendix:cot}
In this section, we present the standardized Chain-of-Thought (CoT) templates used in MolBasic for different molecular understanding tasks. These templates guide the model through structured reasoning processes, ensuring consistent and interpretable outputs. For structure comprehension tasks, we apply CoT only during Stage 2 (reasoning training) for three tasks that showed suboptimal performance in Stage 1: atom counting, total bond counting, and formula conversion. The CoT for these tasks primarily constrains the accuracy of intermediate calculation steps, as illustrated in \cref{fig:cot_examples}. For structural property prediction, the reasoning follows a four-step protocol: atom counting, functional group identification, ring system analysis, and property-specific reasoning (\cref{fig:property_cot_examples}). For molecular optimization, the CoT systematically analyzes polar and lipophilic groups, proposes modification strategies, and verifies property improvements (\cref{fig:optimization_cot_examples}). These standardized templates enable the model to develop robust structural understanding while maintaining transparency in its reasoning process.
\newcommand{\smiles}[1]{\texttt{\seqsplit{#1}}}

\newtcolorbox{cotexample}[1][]{
  enhanced,
  breakable,
  colback=gray!5,
  colframe=gray!50,
  fonttitle=\bfseries,
  title=#1,
  left=2mm,
  right=2mm,
  top=1mm,
  bottom=1mm,
  boxrule=0.5pt,
  before skip=6pt,
  after skip=6pt,
  fontupper=\small
}


\begin{cotexample}[Task: Total Bond Counting]
\textbf{Instruction:} Count the total number of chemical bonds in the molecular structure.

\textbf{Input:} \smiles{C=C(C(=O)O)C1CCC2(O)C=CC(=O)C(C)C2(C)C1}

\textbf{Chain-of-Thought:}
\begin{itemize}[leftmargin=*, nosep]
    \item Parsed the SMILES into atoms and bonds.
    \item Identified two carbonyl groups and all substituent connections.
    \item Enumerated all edges: total = 20.
    \item Verification: $V - 1 + \text{cycles} = 19 - 1 + 2 = 20$ \checkmark.
\end{itemize}

\textbf{Answer:} 20
\end{cotexample}

\begin{cotexample}[Task: Heavy Atom Counting]
\textbf{Instruction:} Count the total number of heavy atoms (non-hydrogen) in the molecule.

\textbf{Input:} \smiles{CC(C)=CCOc1c(C)c(C)c(C(=O)O)c(O)c1C}

\textbf{Chain-of-Thought:}
\begin{itemize}[leftmargin=*, nosep]
    \item First aromatic ring \smiles{c1ccc...c1}: 6 ring carbons = 6 atoms.
    \item Aliphatic chain and branches: 9 C = 9 atoms.
    \item Oxygen atoms, including carbonyl and ether groups: 4 O = 4 atoms.
    \item Total = $6 + 9 + 4 = 19$ heavy atoms.
\end{itemize}

\textbf{Answer:} 19
\end{cotexample}

\begin{cotexample}[Task: Formula Conversion]
\textbf{Instruction:} Generate the standard molecular formula for the molecule.

\textbf{Input:} \smiles{CC(C)=CCOc1cc(=O)oc2cccc(C)c12}

\textbf{Chain-of-Thought:}
\begin{itemize}[leftmargin=*, nosep]
    \item Count carbons from SMILES: 15 total.
    \item Count heteroatoms: O = 3.
    \item Determine hydrogens from substitution and valence:
    \begin{itemize}[leftmargin=*, nosep]
        \item Ring hydrogens: 4 H.
        \item Aliphatic groups: 12 H.
        \item Total H = 16.
    \end{itemize}
\end{itemize}

\textbf{Answer:} C$_{15}$H$_{16}$O$_3$
\end{cotexample}

\begin{center}
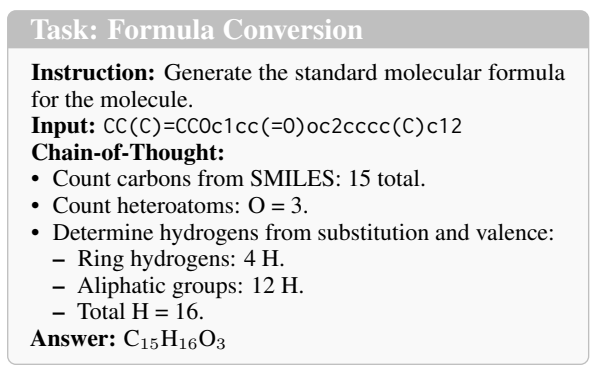

\captionof{figure}{Examples of standardized Chain-of-Thought reasoning for structure perception tasks.}
\label{fig:cot_examples}
\end{center}


\begin{cotexample}[Task: Structural Property Prediction (Example 1)]
\textbf{Input:} \smiles{CC1=CC(=C(C=C1)C)C(=O)CC2(C3=C(C=CC(=C3)Cl)NC2=O)O}

\textbf{Chain-of-Thought:}

\textit{1. Atoms and counts} (Molecular formula: C$_{18}$H$_{16}$ClNO$_3$):
\begin{itemize}[leftmargin=*, nosep]
    \item C: 18, H: 16, N: 1, O: 3, Cl: 1.
    \item Heavy atoms: 23.
\end{itemize}

\textit{2. Functional groups identified}:
\begin{itemize}[leftmargin=*, nosep]
    \item Hydroxyl (--OH), ketone (C=O), amide (--C(=O)NH--), lactam (cyclic amide), and chlorine (--Cl).
\end{itemize}

\textit{3. Ring systems}:
\begin{itemize}[leftmargin=*, nosep]
    \item Total rings: 3, aromatic rings: 2, aliphatic rings: 1, heterocyclic rings: 1.
\end{itemize}

\textit{4. Property reasoning}:
\begin{itemize}[leftmargin=*, nosep]
    \item \textbf{MW}: Sum of atomic weights from C$_{18}$H$_{16}$ClNO$_3$ = 329.78 g/mol.
    \item \textbf{LogP}: Moderately lipophilic due to aromatic rings, balanced by polar groups $\rightarrow$ 3.37.
    \item \textbf{TPSA}: Contributions from N and O atoms $\rightarrow$ 66.4 \AA$^2$.
    \item \textbf{HBD}: 2, from NH/OH groups.
    \item \textbf{HBA}: 3, from N/O atoms with lone pairs.
    \item \textbf{RB}: 3, from non-ring single bonds excluding terminal bonds.
    \item \textbf{QED}: Good drug-likeness $\rightarrow$ 0.85.
\end{itemize}

\textbf{Answer:} MW=329.78, LogP=3.37, TPSA=66.4, HBD=2, HBA=3, RB=3, QED=0.85
\end{cotexample}

\begin{cotexample}[Task: Structural Property Prediction (Example 2)]
\textbf{Input:} \smiles{CCC1=C[N+](=C(C=C1)C)CC2=CC(=O)N3C=CSC3=N2}

\textbf{Chain-of-Thought:}

\textit{1. Atoms and counts} (Molecular formula: C$_{15}$H$_{16}$N$_3$OS$^+$): C: 15, H: 16, N: 3, O: 1, S: 1. Heavy atoms: 20.

\textit{2. Functional groups}: No common functional groups identified.

\textit{3. Ring systems}: Total rings: 3, aromatic rings: 3, heterocyclic rings: 3.

\textit{4. Property reasoning}:
\begin{itemize}[leftmargin=*, nosep]
    \item \textbf{MW}: 286.38 g/mol \quad
    \textbf{LogP}: 1.96, slightly lipophilic \quad
    \textbf{TPSA}: 38.25 \AA$^2$.
    \item \textbf{HBD}: 0 \quad
    \textbf{HBA}: 4 \quad
    \textbf{RB}: 3 \quad
    \textbf{QED}: 0.69.
\end{itemize}

\textbf{Answer:} MW=286.38, LogP=1.96, TPSA=38.25, HBD=0, HBA=4, RB=3, QED=0.69
\end{cotexample}

\begin{center}
\captionof{figure}{Examples of standardized Chain-of-Thought reasoning for structural property prediction.}
\label{fig:property_cot_examples}
\end{center}


\begin{cotexample}[Task: Molecular Optimization (Example 1)]
\textbf{Input:} \smiles{Cc1ccccc1/C=C1/Oc2c(ccc(O)c2C[NH+]2CCC[C@@H](C)C2)C1=O}

\textbf{Chain-of-Thought:}

\textit{1. Original molecule analysis}: C$_{23}$H$_{25}$NO$_3^+$, 4 rings, 2 aromatic rings, LogP: 3.13, HBD: 2, HBA: 3, charge: +1.

\textit{2. Polar groups that decrease LogP}: quaternary ammonium [N+], hydroxyl --OH, ketone C=O, and ether --O--.

\textit{3. Lipophilic groups that increase LogP}: 2 phenyl rings and a 23-carbon skeleton.

\textit{4. Optimization strategy}:
\begin{itemize}[leftmargin=*, nosep]
    \item Neutralize ionic charge, which is expected to provide a major LogP increase.
    \item Reduce HBD: $2 \rightarrow 1$.
    \item Add a piperazine ring with a phenyl substituent.
\end{itemize}

\textit{5. Verification}: LogP: $3.13 \rightarrow 4.64$ ($\Delta = +1.51$) \checkmark.

\textbf{Output:} \smiles{Cc1ccccc1/C=C1\textbackslash Oc2c(ccc(O)c2CN2CCN(c3ccccc3)CC2)C1=O}
\end{cotexample}

\begin{cotexample}[Task: Molecular Optimization (Example 2)]
\textbf{Input:} \smiles{C[NH+](C)C(C)(C)CNC(=O)c1cnn(Cc2ccccc2Cl)c1}

\textbf{Chain-of-Thought:}

\textit{1. Original molecule analysis}: C$_{17}$H$_{24}$ClN$_4$O$^+$, 2 aromatic rings, LogP: 1.24, HBD: 2, HBA: 3, charge: +1.

\textit{2. Polar groups that decrease LogP}: quaternary ammonium [N+], secondary amine --NH--, and amide --CONH--.

\textit{3. Lipophilic groups that increase LogP}: phenyl ring, chlorine, and 17-carbon skeleton.

\textit{4. Optimization strategy}:
\begin{itemize}[leftmargin=*, nosep]
    \item Remove charged quaternary N, which is expected to provide a major LogP increase.
    \item Reduce HBD: $2 \rightarrow 1$.
    \item Add trifluoromethyl --CF$_3$ and phenyl ring.
\end{itemize}

\textit{5. Verification}: LogP: $1.24 \rightarrow 4.86$ ($\Delta = +3.62$) \checkmark.

\textbf{Output:} \smiles{O=C(Nc1ccccc1C(F)(F)F)c1cnn(Cc2ccccc2Cl)c1}
\end{cotexample}

\begin{center}
\captionof{figure}{Examples of standardized Chain-of-Thought reasoning for molecular optimization (LogP improvement).}
\label{fig:optimization_cot_examples}
\end{center}

\section{Use of AI Assistants}
We used AI assistants during the preparation of this work for limited auxiliary purposes, including code debugging and grammar-level writing assistance. All research ideas, experimental design, analysis, conclusions, and final manuscript decisions were made by the authors.

In addition, some of the evaluated baseline systems in our experiments are themselves AI assistants or general-purpose LLMs, such as GPT-series models and Qwen-series models. These models were used only as experimental subjects or baselines for evaluating molecular understanding capabilities, following the experimental protocols described in the paper.

\section{Interpretable Molecular Optimization Examples}
\label{appendix:interpretability}
\begin{figure*}[t]
\centering
\includegraphics[width=2\columnwidth]{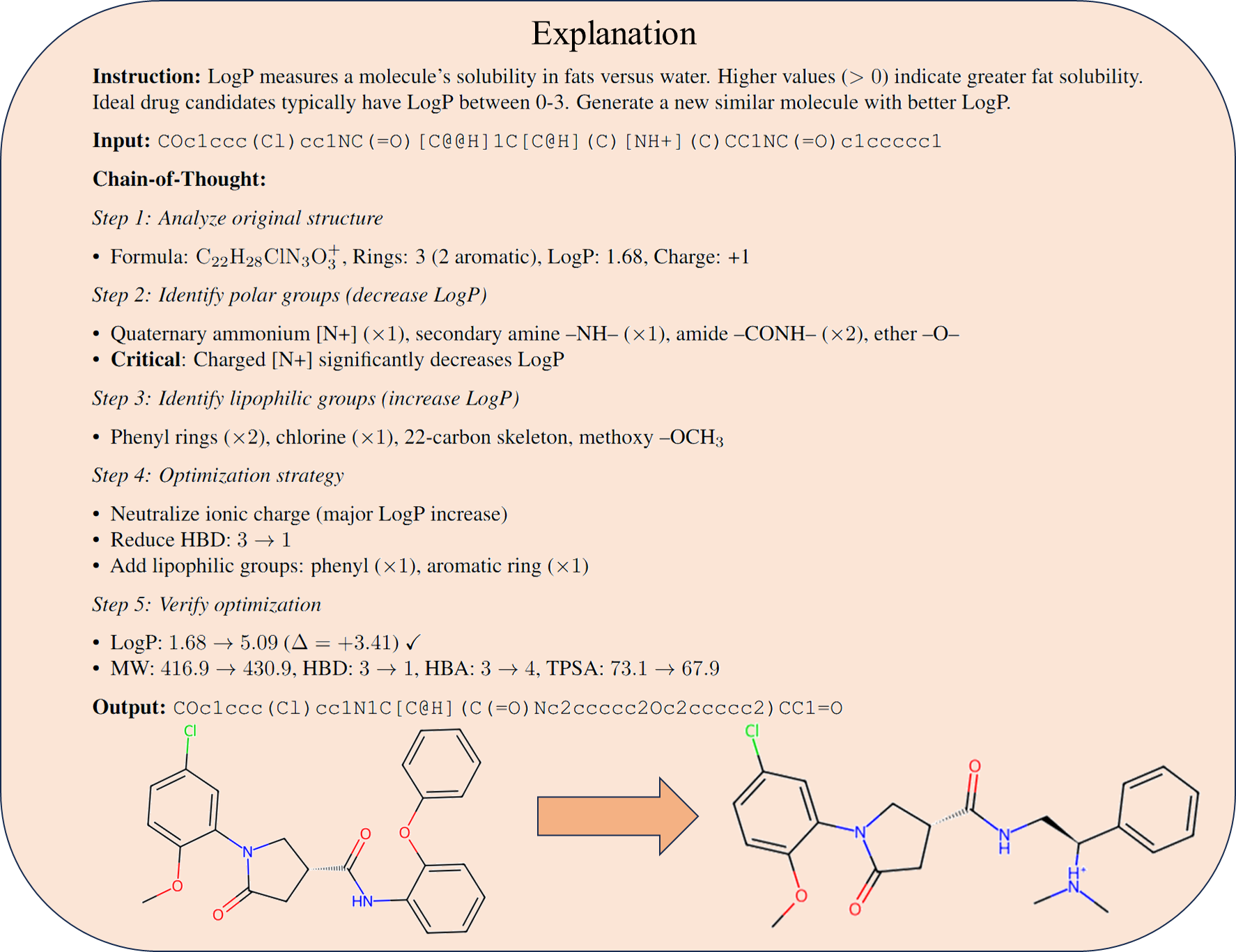}
\caption{Case study 1: Molecular optimization examples with CoT explanations.}
\label{fig:case1}
\end{figure*}

\begin{figure*}[t]
\centering
\includegraphics[width=2\columnwidth]{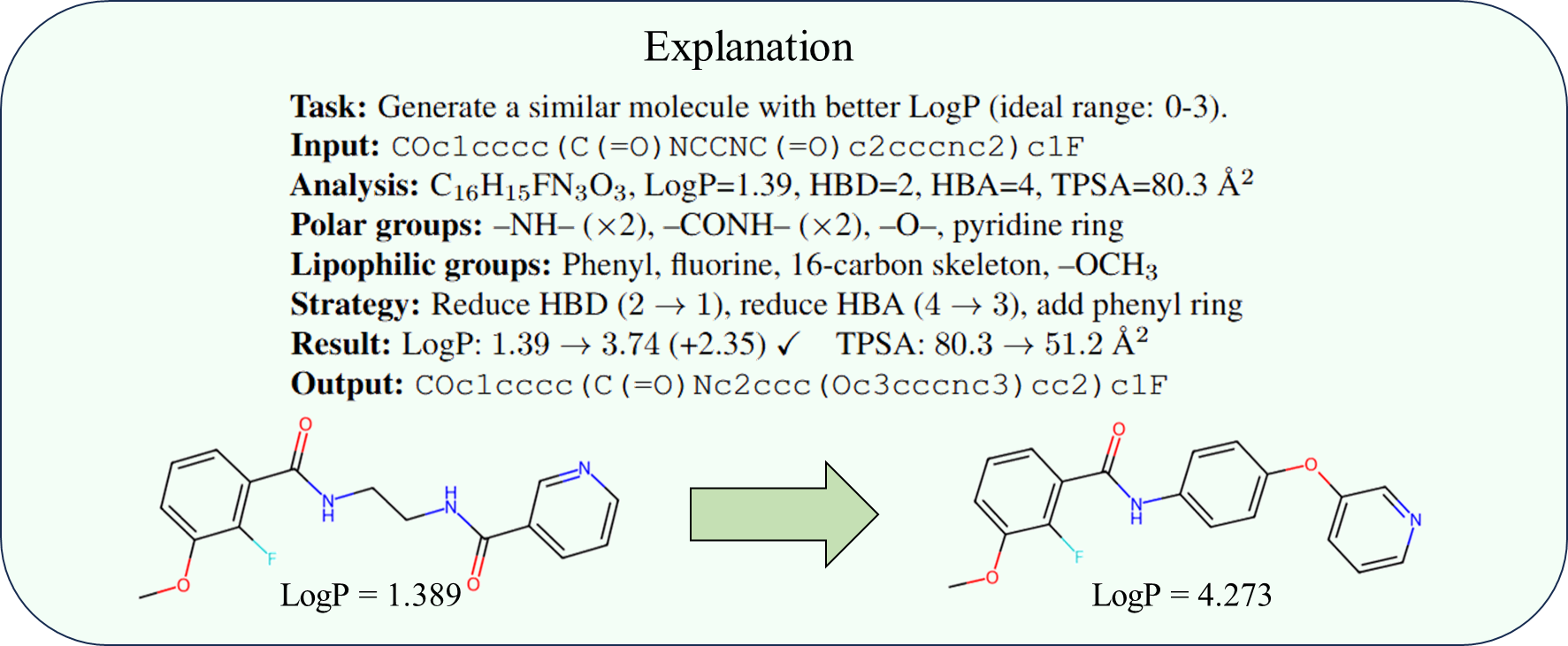}
\vspace{0.2cm}
\includegraphics[width=2\columnwidth]{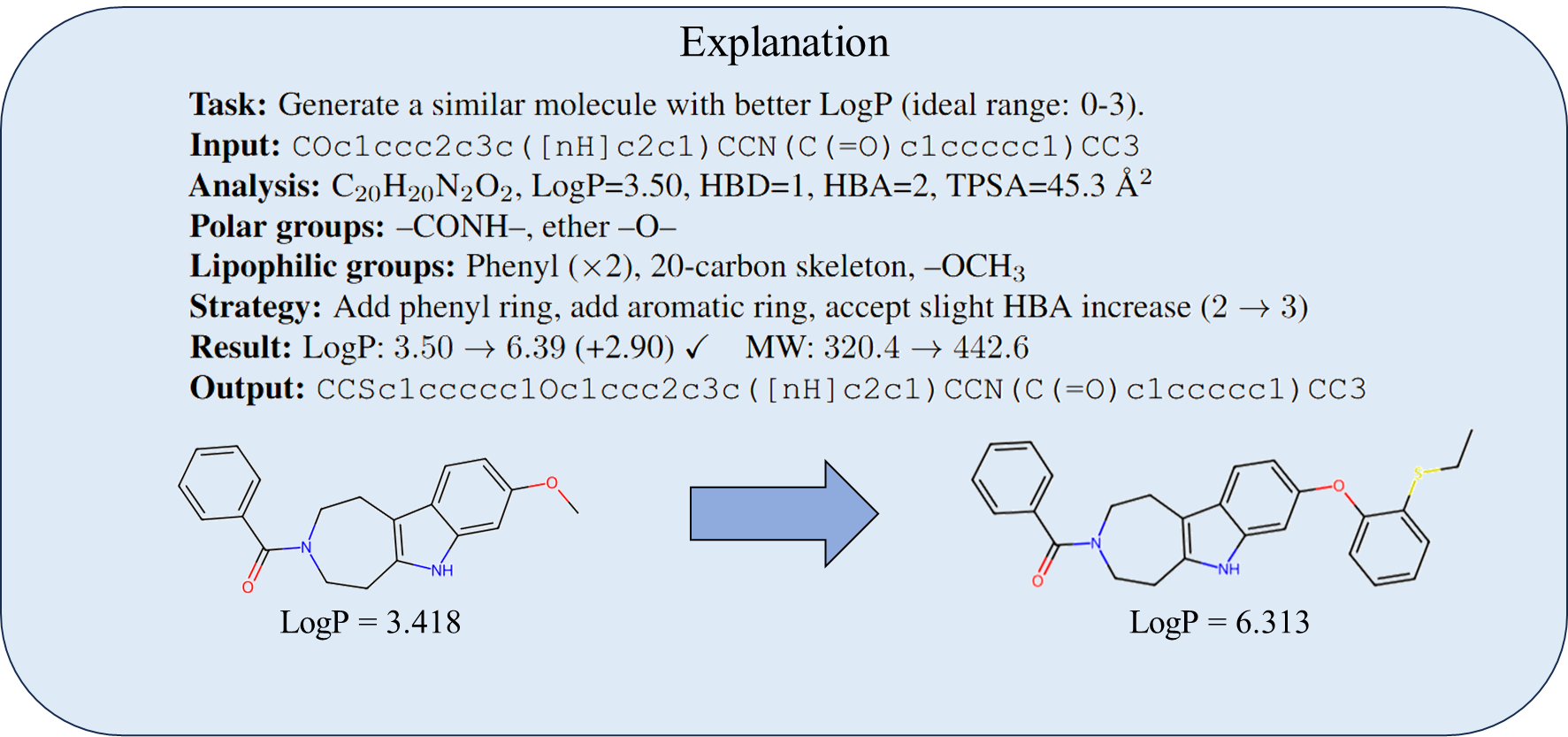}
\caption{Case study 2 and 3: Molecular optimization examples with CoT explanations.}
\label{fig:mol_opt_cases}
\end{figure*}

We present representative examples of MolBasic's responses on the objective molecular optimization task in \cref{4.2.2}. Our framework generates optimized molecules while simultaneously providing interpretable analysis of the optimization rationale, including structural feature identification, modification strategies, and property change verification. \cref{fig:case1,fig:mol_opt_cases} illustrate three such examples with detailed explanations. Note that the intermediate property values in the explanations may not be perfectly accurate, as the model itself exhibits certain prediction errors on property estimation tasks (as shown in \cref{tab:property_prediction}). Nevertheless, the overall trends in property changes remain correct and the errors stay within acceptable ranges, which is sufficient for guiding effective molecular modifications. This demonstrates that MolBasic has developed genuine structural reasoning capabilities for molecular understanding.


\end{document}